\pdfoutput=1

\documentclass[10pt,twocolumn,letterpaper]{article}

\newif\ifreviewsupp
\reviewsuppfalse

\ifreviewsupp\else\usepackage{cvpr}\fi             %
\ifreviewsupp\usepackage[review]{cvpr}\fi    %

\usepackage{pgfplots}
\usepgfplotslibrary{groupplots}
\usepackage{tikz}
\usetikzlibrary{matrix}
\pgfplotsset{compat=1.18}

\usepackage{cuted}
\usepackage{caption}

\preCutedStrip{\vspace{-24pt}}
\postCutedStrip{\vspace{12pt}}

\definecolor{spectral1}{HTML}{9E0142}
\definecolor{spectral2}{HTML}{D53E4F}
\definecolor{spectral3}{HTML}{F46D43}
\definecolor{spectral4}{HTML}{FDAE61}
\definecolor{spectral5}{HTML}{606669} %
\definecolor{spectral6}{HTML}{FFFFBF}
\definecolor{spectral7}{HTML}{E6F598}
\definecolor{spectral8}{HTML}{ABDDA4}
\definecolor{spectral9}{HTML}{66C2A5}
\definecolor{spectral10}{HTML}{3288BD}
\definecolor{spectral11}{HTML}{5E4FA2}

\usepackage[nolist,nohyperlinks]{acronym}[=v2]
\begin{acronym}
\acro{MHSA}[\texttt{MHSA}]{multi-head self-attention}
\acro{LLM}{large language model}
\acro{VLM}{vision language model}
\acro{NLP}{natural language processing}
\acro{RoPE}[\texttt{RoPE}]{rotary positional embeddings}
\acro{CNN}{convolutional neural network}
\acro{ViT}{Vision Transformer}
\acro{MCTP}{Monte Carlo Token Pruning}
\acro{MIM}{masked image modelling}
\acro{SSL}{self-supervised learning}
\acro{MAP}[MAP]{multi-head attention pooling}
\acro{SDPA}[\texttt{SDPA}]{scaled dot-product attention}
\acro{MLP}[\texttt{MLP}]{multi-layer perceptron}
\acro{PCC}{Pearson's Correlation Coefficient}
\end{acronym}

\usepackage[most]{tcolorbox}
\usepackage{xcolor}

\definecolor{boxblue}{RGB}{45,140,210}
\definecolor{boxbluebg}{RGB}{245,249,252}

\definecolor{boxgreen}{RGB}{55,170,75}
\definecolor{boxgreenbg}{RGB}{247,252,247}

\definecolor{boxorange}{RGB}{225,125,30}
\definecolor{boxorangebg}{RGB}{254,248,244}

\newtcolorbox{bluebox}{
    colback=boxbluebg,
    colframe=boxblue,
    boxrule=0.5pt,
    arc=0pt,
    outer arc=0pt,
    left=5pt,
    right=5pt,
    top=4pt,
    bottom=4pt,
    boxsep=0pt,
    before skip=6pt,
    after skip=6pt,
    breakable=true
}

\newtcolorbox{greenbox}{
    colback=boxgreenbg,
    colframe=boxgreen,
    boxrule=0.5pt,
    arc=0pt,
    outer arc=0pt,
    left=5pt,
    right=5pt,
    top=4pt,
    bottom=4pt,
    boxsep=0pt,
    before skip=6pt,
    after skip=6pt,
    breakable=true
}

\newtcolorbox{orangebox}{
    colback=boxorangebg,
    colframe=boxorange,
    boxrule=0.5pt,
    arc=0pt,
    outer arc=0pt,
    left=5pt,
    right=5pt,
    top=4pt,
    bottom=4pt,
    boxsep=0pt,
    before skip=6pt,
    after skip=6pt,
    breakable=true
}

\definecolor{cvprblue}{rgb}{0.21,0.49,0.74}
\usepackage[pagebackref,breaklinks,colorlinks,allcolors=cvprblue]{hyperref}

\def\confName{CVPR}
\def\confYear{2027}

\title{Token Pruning Needs a Protected Sink Token:\\Coherent Spatial Aggregation in Vision Transformers}

\author{Julian Wyatt \qquad James Willoughby \qquad Allison Clement \qquad Irina Voiculescu\\
Department of Computer Science\\
University of Oxford\\
{\tt\small julian.wyatt@cs.ox.ac.uk \quad \url{https://github.com/Julian-Wyatt/MCTP}}
}

\begin{document}

\maketitle
\input{figs/teaser_with_protect_high_norm}
\begin{abstract}
\acfp{ViT} are a predominant architecture for computer vision, with token reduction emerging as a popular paradigm to improve their efficiency.
Prior work compares reduction algorithms, but not how models respond to them, nor how their response changes with model depth.
We propose \acf{MCTP}, which estimates each token's impact on the prediction when tokens are explicitly removed during the forward pass.
Applying \ac{MCTP} to recent pretrained models, we observe that most, such as \mbox{DINOv3}, maintain a similar ranking of token importance at every depth (high \textbf{Coherent Aggregation}), while SigLIP 2 does not.
This coherence requires predictions to survive pruning, which depends on keeping high-norm ``sink'' tokens: prefix tokens are kept by design, while prefix-free SigLIP 2 forms them on patch tokens, which pruning may remove.
We show that in removing DINOv3's prefix tokens, its predictions and coherence collapse, while protecting only 5 tokens for SigLIP 2 cuts the token impact outside the object's bounding box by up to $62\%$.
Beyond this, \ac{MCTP} offers a post-hoc method to evaluate how a pretrained \ac{ViT} responds to pruning, and suggests that token reduction methods should keep ``sinks''.
\end{abstract}
\acresetall

\section{Introduction}
\label{sec:intro}

\acfp{ViT} are the prevalent architecture for image recognition \cite{ViT,touvron2021training,touvron2021going,han2022survey}, vision-language tasks \cite{radford2021learning,zhai2023sigmoid,tschannen2025siglip,cherti2023reproducibleOPENCLIP,tips_v2_paper,bolya2025perception,yang2025visionzip}, dense recognition \cite{ranftl2021visionDPT,strudel2021segmenter,vitadapter,kerssies2025your,amir2021deep} and 3D understanding \cite{wang2025vggt,wang2026vggtomega,chen2026co}. The architecture scales well with both parameter count and dataset size \cite{zhai2022scaling,pmlr-v202-dehghani23a}, leading \acp{ViT} to be the popular candidate for general-purpose feature extraction from frozen or fine-tuned pretrained %
foundation models \cite{caron2021emerging,oquab2023dinov2,simeoni2025dinov3,he2022masked,beit,eva02,tschannen2025siglip,fu2026vision}.

\begin{figure*}[t]
    \centering
    \includegraphics[width=\textwidth]{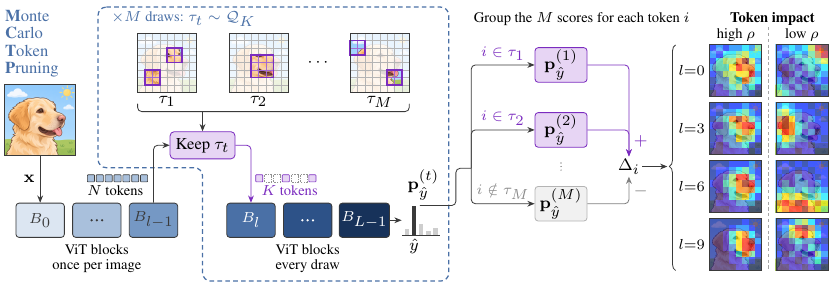}
\caption{\textbf{Monte Carlo Token Pruning.}
Blocks $B_0,\ldots,B_{l-1}$ run once per image. In each of $M$ draws, $\mathcal{Q}_K$ samples a subset $\tau_t$ of the patch tokens, and blocks $B_l,\ldots,B_{L-1}$ run on $\{\mathrm{prefix}\}\cup\tau_t$ to give $\mathbf{p}^{(t)}_{\hat{y}}$. For each token $i$, $\Delta_i$ (\cref{eq:prob-delta}) compares $\mathbf{p}^{(t)}_{\hat{y}}$ over the draws that keep $i$ with those that remove it, giving one token impact map per pruning depth $l$. The mean Spearman $\rho$ over all pairs of these maps measures \textbf{Coherent Aggregation} (\cref{sec:coherent-aggregation}): high $\rho$ when a model's maps agree across depths (left), low $\rho$ when they do not (right).}
    \label{fig:method}
\end{figure*}

A core component of the \ac{ViT} which contributes to its strong representational capacity is the use of \ac{MHSA} to consolidate context across all image tokens; however, this comes at the cost of quadratic computational complexity in the number of image tokens, $N$: $\mathcal{O}(N^2)$. The computational overhead resulting from this quadratic scaling is significant and as such, a substantial subfield of research has emerged exploring methods to reduce the number of tokens and thereby mitigate this impact. These methodological paradigms include: token pruning \cite{rao2021dynamicvit,EViT,kong2022spvit,CROPR,Schlesinger_2026_CVPR,yang2025visionzip,WyaJul_How_MICCAISAT2026,fayyaz2022adaptive}, whereby tokens are identified to be dropped from the sequence, token merging \cite{bolya2023ToMe,bolya2023ToMeDiff,marin2023token,kim2024token,liang2022expediting,chen2026co,zong2022self,Renggli2022LearningTM}, where similar tokens are combined into one, and adaptive patch sizing \cite{APT,kim2026ddit,nakayama2026level}, which allocates different token budgets to different regions of an image. These methods vary substantially in their pruning criteria, as well as the transformer depths at which these decisions are made.

Prior work \cite{haurum2023tokens} systematically investigates token reduction methods, exploring how the performance of different methods varies when changing the keep rates, DeiT \cite{touvron2021training} backbone capacities, and datasets. They primarily examine the token reduction algorithms themselves but do not explore how the response to token reduction may vary between existing pretrained models. This leaves a number of open questions as to the kind of differences pretrained \acp{ViT} may exhibit under token reduction, such as which tokens are important and how consistent that stays throughout the depth of the model.

To investigate the impact of token reduction on various pretrained \acp{ViT}, we introduce \acf{MCTP}: a framework which prunes random token subsets at multiple depths and measures the effect on the prediction. This allows us to estimate not only which regions are important but also how that importance varies throughout the model depth.

Under our \ac{MCTP} framework, we show that most pretrained models remain consistent in which regions carry the largest impact under token removal throughout the full transformer depth, see \cref{fig:teaser}. We term this \textbf{Coherent Aggregation}. For these models, the regions with the largest impact also fall mostly within the object's bounding box, showing high \ac{MCTP} object alignment.

However, SigLIP 2 exhibits low \textbf{Coherent Aggregation} and low \ac{MCTP} object alignment. Unlike the other models, it has no prefix tokens, so ``sink'' tokens \cite{darcet2024vision,jiang2026vision,shi2026vision,xiao2024efficient}, high-norm tokens that hold global image context, form on patch tokens that pruning can remove. Protecting the 5 highest-norm tokens at the point of pruning improves both its \textbf{Coherent Aggregation} and \ac{MCTP} object alignment.

\noindent\textbf{Contributions.} To summarise, our contributions are:
\begin{itemize}
\item We propose \acf{MCTP} to estimate a model's token impact under pruning at varying transformer depths (\cref{sec:MCTP-base}).
\item We observe that most pretrained models remain consistent in which regions they find impactful under token pruning throughout the full transformer depth: high \textbf{Coherent Aggregation} (\cref{sec:coherent-aggregation}).
\item We find that classes with a larger spread of token impact give more stable predictions under pruning, and that coherent models concentrate this impact on the object: high \ac{MCTP} object alignment (\cref{sec:effective-support}).
\item We show that pruning ``sink'' tokens, cause predictions to collapse, and that protecting SigLIP 2's highest-norm tokens at the point of pruning improves its \ac{MCTP} object alignment and prediction stability (\cref{sec:coherent-causes}).
\end{itemize}

\section{Related Work}
\label{sec:related-work}

Our work draws on four areas of research: efficient transformers, token pruning, perturbation-based attribution, and the high-norm ``sink'' tokens that form inside \acp{ViT}.

\textbf{Efficient Transformers.} With the widespread application of \acp{ViT}, improving their efficiency is a popular research domain \cite{JMLR:v20:18-598NeuralArchitectureSearch, liu2022neural,shim2021layer, he2024matters,ma2024era,kitaev2020reformer,katharopoulos2020transformers}. In this work, we focus primarily on efficiency through token pruning.

\textbf{Token Pruning.} Token pruning methods differ in the signal they use to select which tokens to keep. Attention-based methods utilise attention maps or scores to guide token removal \cite{Schlesinger_2026_CVPR,EViT,yang2025visionzip,fayyaz2022adaptive} and learned predictors are trained to make the pruning decision \cite{rao2021dynamicvit,CROPR}. They also prune at different depths, from several blocks throughout the encoder \cite{EViT,rao2021dynamicvit,CROPR,WyaJul_How_MICCAISAT2026} to once after it \cite{yang2025visionzip}. Some, such as ATS \cite{fayyaz2022adaptive} and VisionZip \cite{yang2025visionzip}, can also prune without any further training, so the pretrained model is never adapted to the tokens they remove. \citet{haurum2023tokens} compare these methods with each other, while we study how pretrained models respond to token removal at each depth.

\textbf{Perturbation-based Attribution.} Perturbation methods score image regions by how the prediction changes when they are masked \cite{covert2021explaining}, often by Monte Carlo sampling. RISE \cite{Petsiuk2018rise} weights random masks by the model's output on each masked image. 

For transformers, ViT-Shapley \cite{covert2023learning} masks patches through attention from the input onwards, so removes them only at the input, and trains an explainer to estimate their Shapley values. Most \ac{ViT} attribution methods build maps from attention, patch embeddings or token norms without removing tokens \cite{abnar2020quantifying,chefer2021transformer,xie2023vitcx,wu2024tokentm}. Closest to our work, \citet{wang2026token} remove single visual tokens inside the LLM component from a \ac{VLM} and measure the change in its output; we extend this to Monte Carlo-sampled regions of tokens removed inside the \ac{ViT} encoder. Saliency maps are commonly evaluated by their rank agreement under perturbation \cite{ghorbani2019interpretation,adebayo2018sanity}, and by how much of their mass falls on annotated objects \cite{kohlbrenner2020towards,arras2022clevrxai,shi2026vision}. 

\textbf{Sink Tokens.} StreamingLLM \cite{xiao2024efficient} established that models disproportionately focus on the first tokens in a sequence, which they term ``attention sinks''. In a similar vein, \citet{darcet2024vision} find that artefacts form in the attention maps of modern \acp{ViT}, caused by high-norm outlier tokens on redundant background patches, at positions that vary between images. They remove these outliers by adding learned register \textit{prefix} tokens that take over the high-norm context from the background patches. \citet{jiang2026vision} find that the high-norm tokens appear approximately at block 6 and are caused by highly activating register neurons in intermediate \ac{MLP} layers. At test time, they move these activations into an added register by zeroing the neurons at patch tokens and writing their maximum activation to the register. \citet{shi2026vision} argue that these high-norm tokens are one symptom of lazy aggregation, where image-level supervision leads \acp{ViT} to encode global semantics in background patches, which registers alone do not fully address. Token pruning keeps \textit{prefix} tokens such as [CLS] or registers by design \cite{EViT,rao2021dynamicvit, cao2026putting}, much like the initial [BOS] token in \acp{LLM} \cite{xiao2024efficient}; without them, the sinks can only form on these background patches. Pruning methods do not target sink tokens explicitly, though attention-based or learned scores may keep them implicitly, as sinks draw most of the attention \cite{yang2025visionzip}.

\section{Methodology}

\subsection{Preliminary}

\textbf{\Acfp{ViT}.} Pretrained \acp{ViT} \cite{ViT} take an input image $\mathbf{x} \in \mathbb{R}^{H \times W \times 3}$ and split it into $N$ non-overlapping patches of shape ($H_p$, $W_p$). These patches are then linearly projected into a set of tokens $\mathbf{z}^0 \in \mathbb{R}^{N\times D}$ and prepended with any prefix tokens, $\{\mathrm{prefix}\}$, such as [CLS] and registers. We denote the $i$-th token at transformer block $B_l$ by $\mathbf{z}_{i}^{l}$. The tokens are then processed by $L$ transformer blocks \cite{vaswani2017attention}. For classification, the final representation $\mathbf{z}^L$ provides a single prediction readout by concatenating the [CLS] token $\mathbf{z}_{\text{cls}}$ with the global average pooling (GAP) of the patch tokens $\mathbf{z}_{\text{patch}}$:
\begin{equation}\label{eq:readout}
\mathbf{z}_\text{out} = \text{cat}(\mathbf{z}_{\text{cls}}, \text{pool}(\mathbf{z}_{\text{patch}})).
\end{equation}
For SigLIP 2 \cite{tschannen2025siglip}, which has no [CLS] token, the \ac{MAP} head \cite{lee2019set} output replaces $\mathbf{z}_{\text{cls}}$ in \cref{eq:readout}, keeping the readout dimension ($2D$) the same as for the other models.
The resulting representation $\mathbf{z}_\text{out}$ is projected to $C$ class logits using $W_{\text{probe}} \in \mathbb{R}^{C \times 2D}$ and $\mathbf{b}_{\text{probe}} \in \mathbb{R}^{C}$ and converted to a probability distribution, $\mathbf{p} = \text{softmax}(W_{\text{probe}}\,\mathbf{z}_\text{out} + \mathbf{b}_{\text{probe}})$, with prediction $\hat{y} = \arg\max \mathbf{p}$.
$W_{\text{probe}}$ denotes the linear classifier trained on ImageNet-1k \cite{deng2009imagenet} using the DINOv3 linear probing protocol \cite{simeoni2025dinov3}, without any token pruning. The best probe from the protocol’s hyperparameter sweep, by top-1 accuracy, is then used to evaluate token impact in the following section.

 \subsection{\texorpdfstring{\acf{MCTP}}{Monte Carlo Token Pruning (MCTP)}}\label{sec:MCTP-base}

To assess the post-hoc impact of individual patch tokens on a classification prediction, we introduce \acf{MCTP} (\cref{fig:method}).

Consider keeping only a small subset of tokens $\tau \subset \mathbf{z}^l$ at the start of block $B_l$, where $\tau \sim \mathcal{Q}_K(\mathbf{z}_1,\dots,\mathbf{z}_N)$, $|\tau|{\approx}K$ and $\mathcal{Q}_K$ denotes the random sampling algorithm of $K$ patch tokens where $K \ll N$. Blocks $B_l,\dots,B_{L-1}$ then operate only on the reduced token-stream $\{\mathrm{prefix}\} \cup \tau$. Assuming the original, unpruned prediction from the classification head is the reference prediction $\hat{y}$, then a single ``draw'' of this allows us to determine whether that pruning selection preserves the prediction. We measure the stability, $\hat{s}$, of keeping subset $\tau$ as:
\begin{equation}\label{eq:stability-single-prune}
    \hat{s}(\mathbf{x},\tau) = \mathbf{1}[\arg\max \mathbf{p}(\mathbf{x},\tau,l) = \hat{y}],
\end{equation}
where $\mathbf{1}$ is the indicator function and the readout pooling of patch tokens $\mathbf{z}_{\text{patch}}$ is taken over only the kept tokens $\tau$.

\textbf{Token Impact.} Over $M$ independent Monte Carlo draws of $\tau_1,\dots,\tau_t,\dots,  \tau_M$, the token impact, $\Delta_i$ for the primary predicted class $\hat{y}$ is: 
\begin{equation}
    \Delta_i = \mathbb{E}[\mathbf{p}_{\hat{y}}(\mathbf{x},\tau,l) | i \in \tau] - \mathbb{E}[\mathbf{p}_{\hat{y}}(\mathbf{x},\tau,l) | i \notin \tau], \label{eq:prob-delta}
\end{equation}
\begin{equation}
    \mathbb{E}[\mathbf{p}_{\hat{y}}(\mathbf{x},\tau,l) | i \in \tau] \approx \frac{\sum^M_{t=1} \mathbf{1}[i \in \tau_t]\mathbf{p}_{\hat{y}}(\mathbf{x},\tau_t,l)}{\sum^M_{t=1} \mathbf{1}[i \in \tau_t]}
\end{equation}
where the expectation is over $\tau \sim \mathcal{Q}_K$ (omitted for brevity), and the second term is estimated in the same way over the draws that remove $i$. A larger $\Delta_i$ means $\mathbf{p}_{\hat{y}}$ is higher, on average, when token $i$ is kept than when it is removed. This produces the token impact maps shown in \cref{fig:teaser}. Like Data Banzhaf \cite{wang2023data}, \cref{eq:prob-delta} compares random subsets that keep token $i$ with those that remove it, a comparison they show is more robust to noise in the model’s output than Shapley values.
Unlike RISE \cite{Petsiuk2018rise} and other removal-based explanations \cite{covert2021explaining}, which mask the input, \ac{MCTP} removes tokens inside the network at block $B_l$, and at $l{=}0$ is comparable to the attention masking of ViT-Shapley \cite{covert2023learning}.

In practice, $\mathcal{Q}_K$ samples spatially contiguous regions (square blocks of the patch grid), as uniformly sampled tokens generally change the prediction too little (Supp.~\cref{fig:mctp-sampler-ablation}). We keep $K$ at $10\%$ of the $N$ patch tokens and use $M{=}1{,}000$ draws; Supp.~\cref{sec:MCTP-extra-details} gives details of the sampler and ablates both region size and $M$. As we cache $\mathbf{z}^l$ at block $B_l$ and each draw runs on only 10\% of the tokens, a 1,000-draw run at $l{=}6$ costs about 47 full forward passes, against 1,000 for input masking.

\section{Analysis} \label{sec:analysis} %

\subsection{Coherent Aggregation}\label{sec:coherent-aggregation}

\begin{bluebox}
\textbf{Q:} Does the ranking of token importance stay the same when pruning at different depths?
\end{bluebox}

\textbf{Experimental Setup.} We run \ac{MCTP} over the trained ViT-B/16 linear probes from DINOv3 \cite{simeoni2025dinov3} and SigLIP 2 \cite{tschannen2025siglip}, with four additional backbones \cite{fu2026vision,touvron2021training,tips_v2_paper,eva02} and two other DINOv3 sizes in \cref{tab:backbone-summary} and Supp.~\cref{sec:extended-analysis}. We chose these backbones to span the main pretraining objectives: self-distillation (DINOv3 \cite{simeoni2025dinov3}, and LingBot-Vision \cite{fu2026vision} with dense-feature improvements), supervised classification (DeiT \cite{touvron2021training}), masked image modelling (EVA-02 \cite{eva02}) and image-text pretraining (TIPSv2 \cite{tips_v2_paper} and SigLIP 2 \cite{tschannen2025siglip}), with DINOv3 S/16 and L/16 varying model capacity. The protocol runs over the full $50{,}000$-image ImageNet-1k \cite{deng2009imagenet} validation set at $336\times336$ ($N{=}441$), at pruning depths $l\in\{0,3,6,9\}$. By default, we use the same seed, and so the same draws $\tau$, at every depth, which removes differences in $\tau$ from the depth-wise correlations of token impact and makes them best-case estimates.

\textbf{Spearman Correlation.} To measure the depth-wise correlation, we compute Spearman's rank correlation coefficient \cite{spearman1904proof,ghorbani2019interpretation,adebayo2018sanity} between the token impact maps at each pair of the four depths. For each image, the Spearman Rank Correlation between token impacts at variable pruning depths $l$ quantifies whether tokens retain a similar order of token impact. This value is then averaged across the validation set. The mean Spearman across all pairwise depths is $\rho$ throughout the rest of this paper.

\begin{figure}
    \centering
    \includegraphics[width=\linewidth]{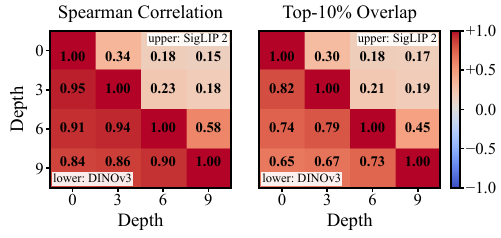}
    \caption{\textbf{Cross-depth consistency.} Signed Spearman correlation (left) and top-10\% overlap (right) of token impacts $\Delta_i$ between pruning depths, for DINOv3 (lower triangle) and SigLIP 2 (upper triangle), averaged over the ImageNet-1k validation set.}
    \label{fig:cross-depth-stability-spearman-main}
\end{figure}

\textbf{Top-10\% Overlap.} We rank tokens by their impact $\Delta_i$ at each depth and, for each pair of \ac{MCTP} pruning depths, report the fraction of tokens shared by their top $10\%$ \cite{ghorbani2019interpretation}, where random rankings would score $0.1$. While the rank correlation compares the full ordering, predictions typically rest on a small set of the most influential tokens, so this assesses whether those regions remain consistent across depths.

\textbf{Prediction Agreement.} To test whether the token impacts directly link to the predictions when pruning at later depths, we keep only the top $25\%$ and $50\%$ of depth-0 $\Delta_i$ tokens, and evaluate the linear probe model prediction agreement using that selection as the core pruning decision over all pruning depths. Formally this is:
\begin{equation}\label{eq:agreement}
    \hat{s}_{\text{prune}} = \frac{1}{|\mathcal{D}|}\sum_{\mathbf{x}\in\mathcal{D}} \mathbf{1}[\arg\max \mathbf{p}(\mathbf{x},\tau^{0}_{\text{top}},l) = \hat{y}],
\end{equation}
where $\tau^0_{\text{top}}$ is the top 25\% or 50\% of tokens by depth-0 $\Delta_i$ and $\mathcal{D}$ is the dataset. As a reference, we also measure this agreement when a uniform random selection of the same number of tokens is kept instead of $\tau^0_{\text{top}}$.

\begin{figure}
\centering
\begin{tikzpicture}
\pgfplotsset{
  impact/.style={
    color=#1, solid, line width=0.8pt,
    mark=x, mark size=1.3pt,
    mark options={solid, fill=#1, draw=#1},
  },
  random/.style={
    color=#1, densely dashed, line width=0.8pt, mark=none,
  },
  decision agreement axis/.style={
    scale only axis=true,
    width=0.37\columnwidth,
    height=0.24\columnwidth,
    xmin=-0.2, xmax=11.2,
    ymin=0, ymax=1.02,
    xtick={0,3,6,9,11},
    ytick={0,0.25,0.5,0.75,1},
    yticklabels={0,0.25,0.5,0.75,1},
    enlarge x limits=false,
    axis x line*=bottom,
    axis y line*=left,
    tick align=outside,
    xmajorgrids, ymajorgrids,
    grid style={gray!25},
    every x tick/.style={color=black, thin},
    every y tick/.style={color=black, thin},
    tick label style={font=\scriptsize},
    title style={font=\scriptsize, yshift=-0.3em},
    ylabel near ticks,
    ylabel shift=-4pt,
  },
}

\begin{groupplot}[
  group style={
    group size=2 by 1,
    horizontal sep=0.05\columnwidth,
    y descriptions at=edge left,
  },
  decision agreement axis,
]
\nextgroupplot[
  title={Retain 25\%},
  ylabel={Agreement},
  legend columns=2,
  legend cell align=left,
  legend style={
    at={(1.07,1.0)}, anchor=south,
    yshift=1.3em,
    draw=none, font=\scriptsize,
    column sep=0.6em, row sep=1pt,
  },
]
\addplot[impact=spectral10] table[col sep=comma, header=true, x=depth, y=prune_reference_agreement]
  {figs/graph_data/prediction_agreement/siglip2_signed_depth0_top25_pruning_stability.csv};
\addlegendentry{SigLIP 2, impact}
\addplot[impact=spectral2] table[col sep=comma, header=true, x=depth, y=prune_reference_agreement]
  {figs/graph_data/prediction_agreement/dinov3_signed_depth0_top25_pruning_stability.csv};
\addlegendentry{DINOv3, impact}
\addplot[random=spectral10] table[col sep=comma, header=true, x=depth, y=prune_reference_agreement]
  {figs/graph_data/prediction_agreement/siglip2_uniform_random_keep25_pruning_stability.csv};
\addlegendentry{SigLIP 2, random}
\addplot[random=spectral2] table[col sep=comma, header=true, x=depth, y=prune_reference_agreement]
  {figs/graph_data/prediction_agreement/dinov3_uniform_random_keep25_pruning_stability.csv};
\addlegendentry{DINOv3, random}

\nextgroupplot[title={Retain 50\%}, yticklabels={}, ylabel={}]
\addplot[impact=spectral10] table[col sep=comma, header=true, x=depth, y=prune_reference_agreement]
  {figs/graph_data/prediction_agreement/siglip2_signed_depth0_top50_pruning_stability.csv};
\addplot[impact=spectral2] table[col sep=comma, header=true, x=depth, y=prune_reference_agreement]
  {figs/graph_data/prediction_agreement/dinov3_signed_depth0_top50_pruning_stability.csv};
\addplot[random=spectral10] table[col sep=comma, header=true, x=depth, y=prune_reference_agreement]
  {figs/graph_data/prediction_agreement/siglip2_uniform_random_keep50_pruning_stability.csv};
\addplot[random=spectral2] table[col sep=comma, header=true, x=depth, y=prune_reference_agreement]
  {figs/graph_data/prediction_agreement/dinov3_uniform_random_keep50_pruning_stability.csv};
\end{groupplot}

\node[anchor=north, font=\footnotesize]
  at ($(group c1r1.south)!0.5!(group c2r1.south) + (0,-1.5em)$)
  {Pruning depth $l$};
\end{tikzpicture}
\caption{\textbf{Prediction agreement.} Agreement $\hat{s}_\text{prune}$ (\cref{eq:agreement}) when pruning the linear-probed models at depth $l$ using the highest depth-0 impacts $\Delta_i$ (solid) versus uniform random selection (dashed), retaining 25\% (left) and 50\% (right) of tokens, over the ImageNet-1k validation set.}
\label{fig:pruning-decision-agreement}
\end{figure}
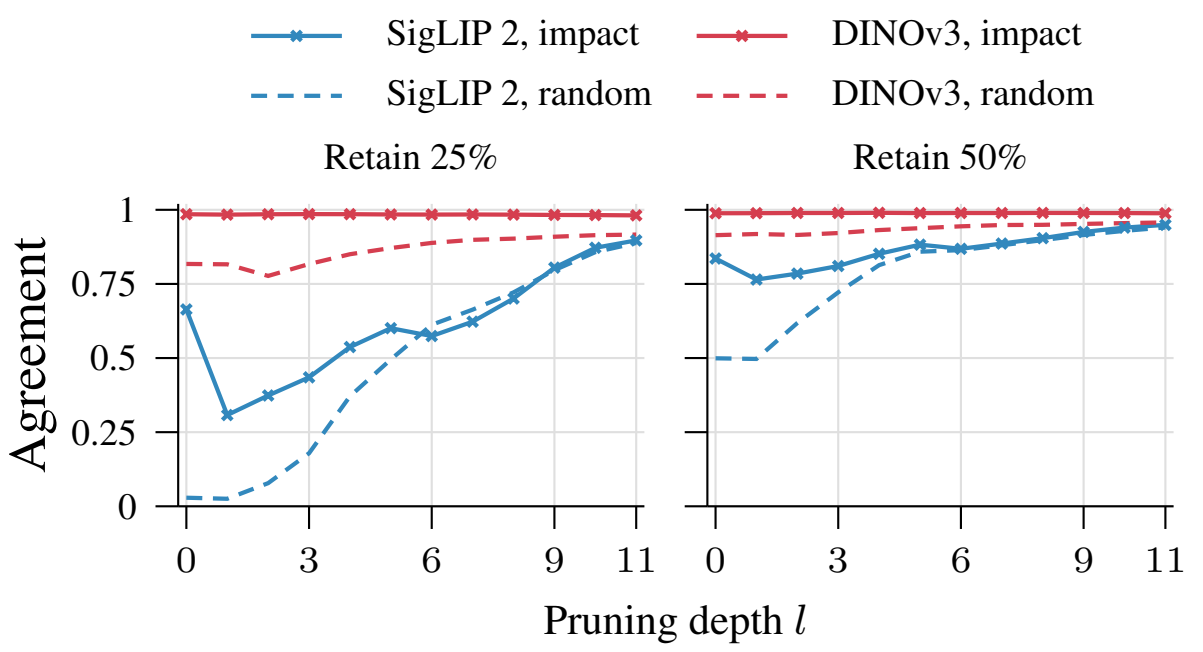

\paragraph{Experiment Results.} 
The results of these experiments are shown in \cref{fig:cross-depth-stability-spearman-main,fig:pruning-decision-agreement}, with additional model comparisons in Supp.~\cref{sec:extended-analysis}. Across each measure, DINOv3 exhibits considerably stronger cross-depth consistency than SigLIP 2. The rank correlation and top-$10\%$ overlap measures in \cref{fig:cross-depth-stability-spearman-main} show that this consistency holds both for the overall ordering and among the most impactful tokens. \cref{fig:pruning-decision-agreement} further demonstrates that this consistency is reflected in the downstream prediction. DINOv3 retains a strong prediction agreement under uniform random pruning, and when using the depth-0 pruning decision the agreement stays at approximately $98.2\%$ and $98.9\%$ for the top $25\%$ and $50\%$ keep rates respectively over all pruning depths. We provide a multi-seed comparison in Supp.~\cref{tab:multi-seed}, which shows that SigLIP 2's token impacts at the same depth $l$ agree with $\rho$ of $0.56$--$0.88$, while across depths a single seed only agrees at $0.28$ (\cref{tab:backbone-summary}). This indicates that the incoherence is not an artefact of \ac{MCTP} seed noise. Of the eight models we evaluate, SigLIP 2 is the only one with low coherence (the others reach $\rho$ of $0.75$--$0.91$); we explore why in \cref{sec:coherent-causes}.

\begin{bluebox}
\textbf{A:} Yes, the token importance from pretrained models remain consistent throughout the transformer forward pass, exhibiting high \textbf{Coherent Aggregation}; SigLIP 2, however, is the exception. 
\end{bluebox}

\subsection{Effective Support}\label{sec:effective-support}

\begin{greenbox}
\textbf{Q:} How does the evidence concentration of a prediction relate to its robustness under pruning, and is that evidence on the object?
\end{greenbox}

\textbf{Experimental Setup.} We reuse the token impact maps and \ac{MCTP} draws from \cref{sec:coherent-aggregation}. The main text shows DINOv3 \cite{simeoni2025dinov3} and SigLIP 2 \cite{tschannen2025siglip}; the other six models are in \cref{tab:backbone-summary} and Supp.~\cref{sec:extended-analysis}.

\textbf{Effective Support.} Evidence concentration is quantified using positive impact entropy $H^+$, similar to \cite{bhatt2020evaluating} and its effective support $\phi^+$, which illustrates the general spread of image-wise token impacts proportional to the number of patch tokens. Let $i, j\in\{1,\dots,N\}$ index the $N$ patch tokens:
\begin{figure}
    \centering
    \includegraphics[width=\linewidth]{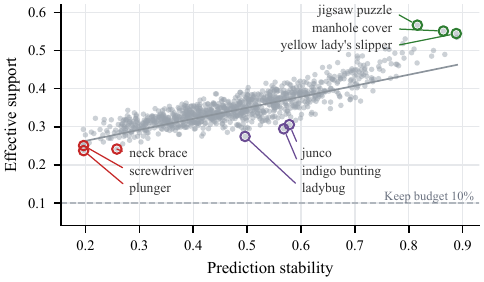}
\caption{\textbf{Effective support against stability.} Class-wise effective support $\phi^+$ (\cref{eq:effective-support}) against prediction stability $\hat{s}(k)$ (\cref{eq:prediction-stability}) for DINOv3 at $l{=}3$, over the ImageNet-1k validation set. Low- and high-support classes are marked in red and green, and the classes furthest below the line of best fit in purple. Pearson correlation 0.86.}
    \label{fig:effective-support-stab-main}
\end{figure}

\begin{gather}
    q_i^+ =
    \frac{\max(\Delta_i,0)}
         {\sum_j \max(\Delta_j,0)},
    \label{eq:positive-impact} \\[1pt]
    H^+ = -\sum_i q_i^+\log q_i^+,
    \qquad
    \phi^+ = \frac{e^{H^+}}{N}.
    \label{eq:effective-support}
\end{gather}
High $\phi^+$ indicates that the impact is shared across a larger fraction of tokens across the image, while low $\phi^+$ indicates that the positive token impact is concentrated in a smaller region in the image.

\textbf{Prediction Stability.} Prediction stability is then obtained by averaging \cref{eq:stability-single-prune} over all \ac{MCTP} draws and images belonging to class $k$:
\begin{equation}\label{eq:prediction-stability}
    \hat{s}(k) = \frac{1}{|\mathcal{D}_k|}\sum_{\mathbf{x}\in\mathcal{D}_k} \frac{1}{M}\sum_{t=1}^{M} \hat{s}(\mathbf{x},\tau_t),
\end{equation}

where $\mathcal{D}_k$ is the set of images from the dataset $\mathcal{D}$ belonging to class $k$, $\hat{s}$ is then the overall average stability. We note that here, $k$ may not necessarily match $\hat{y}$, as we aim to match the unpruned prediction rather than necessarily the correct prediction \cite{hendrycks2019benchmarking,hooker2019benchmark,dutta2024accuracy}. This has the core benefit of isolating stability under pruning against a model's ability to correctly classify an image.
\cref{eq:prediction-stability} measures how reliably sampled token selections preserve the unpruned prediction. Examining its relationship with effective support relates the extent of positive evidence to the sensitivity of a class's prediction over retained tokens.

\textbf{Off-Box Mass.} Lastly, to validate an image's token impact, and thus the verification of effective support, we follow a similar process to \cite{kohlbrenner2020towards,arras2022clevrxai,shi2026vision}. Using the object annotation boxes available for the ImageNet \cite{deng2009imagenet} validation set, we measure the proportion of positive token impact that lies outside of the bounding box. This allows us to quantify \ac{MCTP} object alignment. We select off-box rather than in-box because objects rarely conform exactly to the annotated bounding box: the box gives an estimate of object location, so impact outside it is more reliably off the object, whereas impact inside it may still lie on background. Since \ac{MCTP} considers token impact from the perspective of protecting the unpruned prediction, we only consider the case where $\hat{y} = y$, so that the highlighted impact relates to the bounding box. Consequently, each model's off-box mass is averaged over different validation subsets. We also take the union of multiple boxes. Formally, for normalised positive token impacts $q_i^+$ and the set of tokens $\mathcal{G}$ inside the box, this is
\begin{equation}\label{eq:off-box}
m = \frac{\sum_{i\notin \mathcal{G}} q_i^+}{(N-|\mathcal{G}|)/N},
\end{equation}
where dividing by the fraction of tokens outside the box normalises for box size, to give a mass value over the whole image and not an average per token. Therefore, uniformly spread impact over all tokens gives $m = 1$ and lower values indicate impact concentrated on the object. Examples of these boxes and their masses can be found in Supp.~\cref{fig:off-box-mass-examples}.
\input{figs/effective_support_examples}

\paragraph{Experiment Results.} The results are shown in \cref{fig:effective-support-stab-main,fig:support-examples,tab:backbone-summary}. Class-wise effective support and prediction stability are clearly correlated in \cref{fig:effective-support-stab-main} (Pearson correlation 0.86). Under a 10\% keep budget, broad class evidence survives most sampled token selections, while concentrated evidence survives only when its few tokens are kept, which needs precise token selection but could allow more aggressive pruning. The qualitative examples in \cref{fig:support-examples} suggest that classes occupying more of the image tend to have broader effective support and images from the same class can still have varied effective support. The highlighted classes in \cref{fig:effective-support-stab-main} recur for the other coherent models (Supp.~\cref{fig:effective-support-lingbot-deit,fig:effective-support-eva-tips,fig:effective-support-small-large}): screwdriver stays at the low-support end and jigsaw puzzle at the high end, and junco and other birds are always among the classes furthest below the trend, keeping their prediction more often than their concentrated support suggests.

SigLIP 2's class evidence, however, does not lie on the object (off-box mass $0.94$ and $0.89$ at $l{=}0$ and $9$, against $0.29$--$0.38$ for the other models, \cref{tab:backbone-summary}). SigLIP 2 also has the most concentrated effective support of any model at $l{=}0$ and $3$, and only approaches DINOv3's by $l{=}9$ (Supp.~\cref{fig:effective-support-distributions,fig:depth-wise-siglip-effective-support}). Together with the low cross-depth token-impact correlations in \cref{fig:cross-depth-stability-spearman-main}, these shifts suggest that both the extent and location of SigLIP 2's class evidence change with pruning depth.

\begin{table}[t]
\caption{\textbf{Backbone summary.} All values are over the ImageNet-1k validation set. Top-1 (\%) is unpruned linear probe accuracy using [CLS]+GAP (\cref{eq:readout}). $\rho$ averages per-image Spearman correlations of signed token impacts over all depth pairs (\cref{fig:cross-depth-stability-spearman-main}). Off-box mass $m$ (\cref{eq:off-box}) measures positive impact outside ground-truth-class boxes for correct unpruned predictions; uniform impact gives $m{=}1$. $\hat{s}_{l=6}$ is prediction stability (\cref{eq:prediction-stability}) at $l{=}6$.}
\label{tab:backbone-summary}
\centering
\small
\setlength{\tabcolsep}{3pt}
\begin{tabular*}{\columnwidth}{@{\extracolsep{\fill}}lccccc@{}}
\toprule
 & & & \multicolumn{2}{c}{Off-box $\downarrow$} & \\
\cmidrule(lr){4-5}
Model & Top-1 $\uparrow$ & $\rho\uparrow$
      & $l{=}0$ & $l{=}9$ & $\hat{s}_{l=6}\,(\%)\uparrow$ \\
\midrule
DINOv3 B/16 \cite{simeoni2025dinov3}& 85.3 & 0.90 & 0.31 & 0.31 & 52.03 \\
LingBot-Vision B/16 \cite{fu2026vision} & 84.6 & 0.89 & 0.31 & 0.29 & 45.90 \\
DeiT B/16\textsuperscript{b} \cite{touvron2021training}
    & 82.0 & 0.83 & 0.36 & 0.37 & 50.35 \\
TIPSv2 B/14\textsuperscript{c} \cite{tips_v2_paper}
    & 82.7 & 0.82 & 0.32 & 0.30 & 36.54 \\
EVA-02 B/14\textsuperscript{c} \cite{eva02}
    & 82.7 & 0.75 & 0.35 & 0.38 & 43.23 \\
SigLIP 2 B/16\textsuperscript{a} \cite{tschannen2025siglip}
    & 84.9 & 0.28 & 0.94 & 0.89 & 8.69 \\
\midrule
DINOv3 S/16 \cite{simeoni2025dinov3}& 80.9 & 0.90 & 0.32 & 0.30 & 39.49 \\
DINOv3 L/16\textsuperscript{d} \cite{simeoni2025dinov3}
    & 87.5 & 0.91 & 0.31 & 0.33 & 60.85 \\
\bottomrule
\end{tabular*}
\par\smallskip
\begin{minipage}{\columnwidth}
\footnotesize
\textsuperscript{a}No CLS token; uses the multi-head attention pooling (MAP) output in its place.
\textsuperscript{b}Uses its original ImageNet-1k fine-tuned CLS classifier.
\textsuperscript{c}Patch projection resampled from 14 to 16 pixels.
\textsuperscript{d}Uses $l{=}18$ for the $l{=}9$ off-box column and $l{=}12$ for the $\hat{s}_{l=6}$ column, matching relative network depths.
\end{minipage}
\end{table}

\begin{greenbox}
\textbf{A:} Evidence concentration varies across classes and correlates with stability under pruning: concentrated evidence suggests scope for aggressive pruning with precise token selection, while diffuse evidence indicates a greater robustness to pruning decisions. Further, coherent models put that evidence on the object ($m{\approx}0.3$), while SigLIP 2's does not  ($m{\approx}0.94$).
\end{greenbox}

\subsection{Sink Tokens}\label{sec:coherent-causes}

\begin{orangebox}
\textbf{Q:} What breaks Coherent Aggregation?
\end{orangebox}

SigLIP 2 is a notable outlier in terms of coherence $\rho$, stability $\hat{s}$ and off-box mass $m$ under token pruning in comparison to the other models evaluated (\cref{tab:backbone-summary}). In addition to this, we note that for SigLIP 2 the patch tokens of highest average impact are consistently located in corner regions of the image (Supp.~\cref{fig:global-impact}). SigLIP 2's behaviour is not explained by its pretraining objective alone: TIPSv2, also trained on image-text pairs, remains coherent ($\rho{=}0.82$, \cref{tab:backbone-summary}). What SigLIP 2 alone lacks is prefix tokens, as every other backbone carries at least a [CLS] token. Building on \citet{darcet2024vision}, who showed that high-norm tokens hold global context, we hypothesise that the absence of prefix tokens in SigLIP 2 leads to high-norm sink tokens forming on patch tokens, which are not protected from pruning, rather than on prefix tokens, which are. Pruning these high-norm tokens during \ac{MCTP} then corrupts the representation, resulting in this outlier behaviour.

\textbf{Token Norms.} First, we explore token norms throughout the transformer depth by plotting a log-scale heatmap of $\ell_2$ norms for all patch tokens and, for DINOv3, also the prefix tokens. We scale all norms by the median patch-token norm for each image, model, and block.

\textbf{Prefix Removed.} As an interventional test, we aim to break DINOv3's coherence by running the \ac{MCTP} pipeline and explicitly pruning its protected prefix tokens at block $B_l$. This evaluates whether the high coherence is a property of the prefix tokens. As our core \ac{MCTP} method uses both [CLS] and GAP for readouts (\cref{eq:readout}), in this case we only use GAP.

\input{figs/token_norm_main_paper}

\textbf{Test-Time Registers.} To intervene from the opposing axis, we aim to fix SigLIP 2's coherence. We apply \ac{MCTP} to SigLIP 2 with test-time registers \cite{jiang2026vision} (\cref{sec:related-work}), where \ac{MCTP} always keeps the new register. \textbf{Empty Register.} As a control, we add the same new register without moving the high-norm activations into it.

\textbf{Top Five Protected.} Next, we locate the five highest $\ell_2$-norm tokens at the point of pruning (block $B_l$) and protect these from \ac{MCTP}: $\mathbf{P}_l$. Each \ac{MCTP} draw thus retains the union $\tau \cup \mathbf{P}_l$. Also, because $\mathbf{P}_l$ is protected during \ac{MCTP}, its tokens have an undefined $\Delta$ value and are therefore ignored for results using protected patch tokens. As a better comparison for this test, we provide the baseline (matched) result in \cref{tab:sink-interventions}, which rescores SigLIP 2 without $\mathbf{P}_l$. \textbf{Random Five Protected.} As a control, we repeat this procedure with five patch tokens sampled uniformly before \ac{MCTP}. This selection is fixed per image and over all pruning depths.

\begin{figure}[t]
    \centering
    \sbox0{\includegraphics{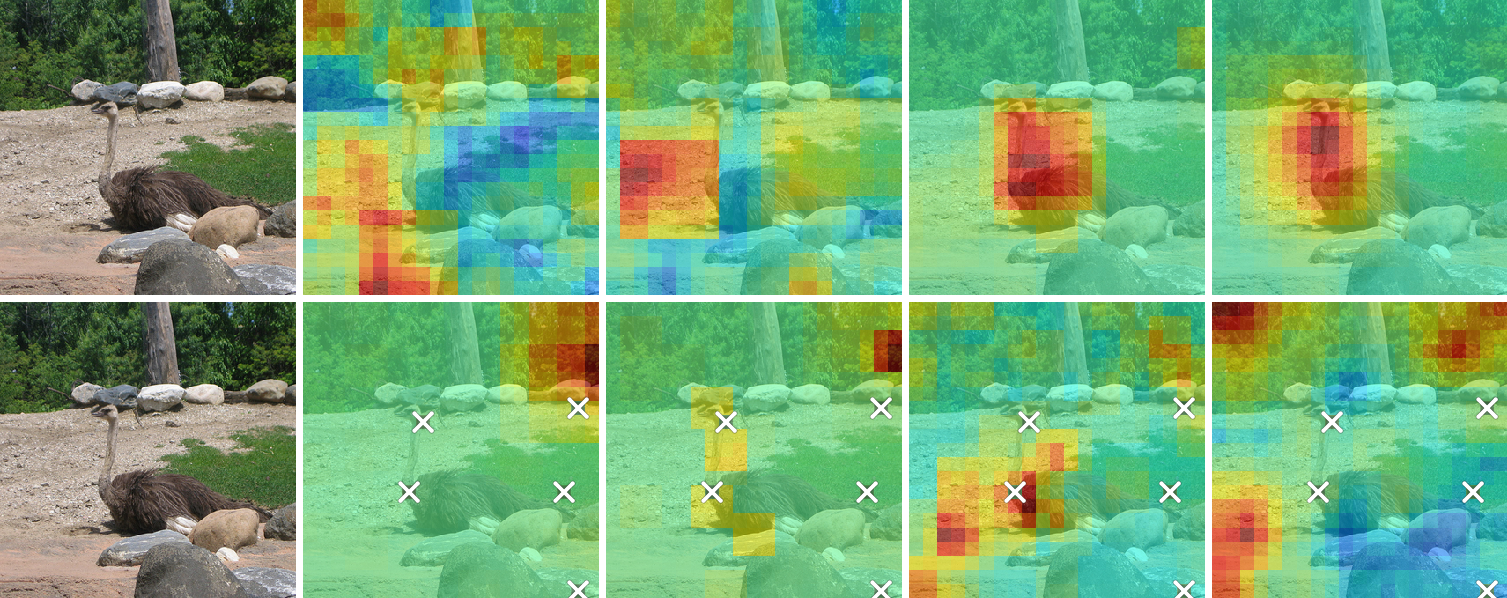}}%
    \edef\imgtrim{\the\dimexpr 0.2\wd0\relax}%
 \begin{tikzpicture}

    \node[inner sep=0, anchor=south west] (image1) at (0,0) {\includegraphics[trim=\imgtrim{} 0pt 0pt 0pt, clip, width=0.8\linewidth]{figs/images/Analysis_3/interventional_token_impacts_DINOv3_siglip2_fix_control_crosses.png}};

    \begin{scope}[x={(image1.south east)}, y={(image1.north west)}]

        \node[rotate=90, anchor=south] at (0,0.75)
            {\footnotesize\shortstack{DINOv3\\[-2pt]No Prefix}};

        \node[rotate=90, anchor=south] at (0,0.25)
            {\footnotesize\shortstack{SigLIP 2\\[-2pt]Control}};

        \node[anchor=south] at (1/8,1) {0};
        \node[anchor=south] at (3/8,1) {3};
        \node[anchor=south] at (5/8,1) {6};
        \node[anchor=south] at (7/8,1) {9};

    \end{scope}

\end{tikzpicture}
\caption{\textbf{Interventional token impacts.} Token impact maps for the input image in \cref{fig:teaser}, row 1, at each pruning depth $l$. \textit{No Prefix} (top): DINOv3 with its prefix tokens removed at the point of pruning. \textit{Control} (bottom): SigLIP 2 with five random tokens (white $\times$) protected from pruning. Compare with \cref{fig:teaser} (right), which protects its five highest-norm tokens.}
    \label{fig:interventional-token-impact}
\end{figure}

\paragraph{Experiment Results.} The results for this set of experiments are found in \cref{fig:token-norms-main,fig:interventional-token-impact,tab:sink-interventions}. Firstly, as seen in \cref{fig:token-norms-main}, the token norms for DINOv3 are located primarily in the register tokens whereas SigLIP 2's appear in a few patch tokens from block 6, matching the OpenCLIP \cite{cherti2023reproducibleOPENCLIP} result from \cite{jiang2026vision}. Furthermore, the specific tokens align to the background regions from Supp.~\cref{fig:global-impact}. However, as highlighted in \cite{jiang2026vision,darcet2024vision}, which specific tokens become high norm varies between images which makes it difficult to know beforehand.

As expected, removing DINOv3's prefix tokens at the point of pruning raises the off-box mass $m$ from $0.30 \to 0.82$ at $l{=}3$ and destroys the model's stability $\hat{s}$, which falls from $49.4\% \to 3.9\%$ at $l{=}6$. Consequently, this also breaks coherence with the mean Spearman $\rho$ falling from $0.90 \to 0.34$. While this is unsurprising, it is interesting that despite their lower norm relative to the prefix tokens, DINOv3's patch tokens appear to have already gathered spatial context from the sinks, as indicated by low off-box mass even with low stability at $l{=}6$. Its token impact maps stay largely on the object when the prefix is removed at $l{=}6$ and $l{=}9$ ($m$ of $0.41$ and $0.29$), whereas SigLIP 2's maps are notably worse without its high-norm tokens.

Meanwhile, SigLIP 2 improves with any targeted token norm intervention.  Specifically, introducing test-time registers improved the off-box mass $m$ from $0.99 \to 0.40$ and stability from $8.7\% \to 30.9\%$ both at $l{=}6$. However, as high-norm outlier tokens typically only appear at the sixth block onwards, this has little effect on pruning at $l{=}3$. The empty register leaves the off-box mass and stability the same as SigLIP 2, so the gain comes from moving the high-norm activations into a token that pruning keeps, rather than naïvely adding the additional token.

Protecting the five highest-norm tokens at the point of pruning improves the off-box mass $m$ from $0.95 \to 0.36$ and stability from $8.7\% \to 36.6\%$ at $l{=}6$, and $m$ from $1.14 \to 0.67$ and stability from $0.6\% \to 10.6\%$ at $l{=}3$ (\cref{tab:sink-interventions} and Supp.~\cref{tab:stability-depth}). It also matches or exceeds test-time registers at every depth, without changing the model's activations. Notably, this is the only experiment that improves the $l{=}3$ mass, yet the tokens it protects there are not the ones that become sinks in the unpruned model: only 3.1\% of the final block's five highest-norm tokens are among the top five at block 3 (only 2\% better than chance), against 80\% at block 6 (Supp.~\cref{fig:topk-norm-overlap}). Instead, they become sinks after pruning: when pruned at $l{=}3$, SigLIP 2 rarely forms a high-norm sink (in only 12--16\% of draws, and 17\% at block 6 with the top five protected). When one does form at block 6, it is one of the protected top five 92\% of the time, against 14\% for five random protected tokens (Supp.~\cref{tab:sink-formation}). We also note that the coherence (mean Spearman) $\rho$ rises from $0.28 \to 0.40$ in the depth pairs where stability recovers: $3\leftrightarrow6$ more than doubles from $0.23 \to 0.50$ and $6\leftrightarrow9$ rises from $0.58 \to 0.70$. However, pairs with $l{=}0$ hold the mean down, as stability at $l{=}0$ only reaches $2.7\%$ (Supp.~\cref{tab:stability-depth}).  Lastly, protecting any five random tokens shows little change to off-box mass $m$, implying that our finding is not an artefact of our \ac{MCTP} method nor the sampling scheme $\mathcal{Q}_K$.

\begin{table}[t]
\caption{\textbf{Sink interventions.} Off-box mass $m$ (\cref{eq:off-box}) at each pruning depth $l$ and prediction stability $\hat{s}_{l=6}$ (\cref{eq:prediction-stability}), over the ImageNet-1k validation set. $\hat{s}$ is measured against the original, unmodified model's unpruned top-1. DINOv3 rows use the GAP probe. Protected tokens are excluded from the map scores, and baseline (matched) excludes the top-five positions from the unmodified SigLIP 2 maps.}
\label{tab:sink-interventions}
\centering
\small
\setlength{\tabcolsep}{3pt}
\begin{tabular}{@{}lccccc@{}}
\toprule
 & \multicolumn{4}{c}{Off-box $\downarrow$} & \\
\cmidrule(lr){2-5}
Intervention & $l{=}0$ & $l{=}3$ & $l{=}6$ & $l{=}9$
             & $\hat{s}_{l=6}\,(\%)\uparrow$ \\
\midrule
DINOv3 B/16
    & 0.29 & 0.30 & 0.30 & 0.29 & 49.36 \\
\quad Prefix removed
    & 0.89 & 0.82 & 0.41 & 0.29 & 3.86 \\
\midrule
SigLIP 2 B/16
    & 0.94 & 1.17 & 0.99 & 0.89 & 8.69 \\
\quad Baseline (matched)
    & 0.93 & 1.14 & 0.95 & 0.87 & 8.69 \\
\quad Top five protected
    & 0.93 & 0.67 & 0.36 & 0.54 & 36.60 \\
\quad Test-time registers 
    & 1.08 & 1.13 & 0.40 & 0.61 & 30.86 \\
\quad Random five protected
    & 1.00 & 1.16 & 1.02 & 0.93 & 11.71 \\
\quad Empty register
    & 1.06 & 1.17 & 0.98 & 0.88 & 8.74 \\
\bottomrule
\end{tabular}
\end{table}

\begin{orangebox}
\textbf{A:} Pruning the high-norm sink tokens that hold global context collapses the prediction, and with it \textbf{Coherent Aggregation}. In most models these include prefix tokens \cite{darcet2024vision}, which pruning protects, but the prefix-free SigLIP~2 forms them on patch tokens, which pruning removes. Protecting SigLIP~2's five highest-norm tokens improves its \ac{MCTP} object alignment and prediction stability and doubles its coherence between depths 3 and 6.
\end{orangebox}

\section{Limitations and Future Work}

\textbf{Limitations.} \ac{MCTP} is too costly to make pruning decisions at inference, so we design it as a post-hoc analysis tool that allows us to explore how pretrained models respond to pruning, with its cost paid only once per model. We do not evaluate existing token pruning methods, nor propose a new pruning method: protecting sink tokens here tests why SigLIP 2 fails under pruning, and its effect within existing methods remains future work. However, what remains uncertain is whether image context is required outside of the high token impact regions under pruning. Of the eight models we evaluate, only SigLIP 2 shows low \textbf{Coherent Aggregation}, so our sink interventions only cover a single prefix-free model; however SigLIP encoders \cite{tschannen2025siglip,zhai2023sigmoid} are the vision encoders widely used in \acp{VLM} such as PaliGemma \cite{beyer2024paligemma,steiner2024paligemma} and LLaVA-OneVision \cite{li2025llavaonevision}. Lastly, protecting tokens cannot help at $l{=}0$, as the model has not yet embedded image context into any patch tokens.

\textbf{Future Work.} We will expand our analysis to cover dense predictions, which is fundamentally more difficult under token reduction \cite{liang2022expediting,tang2023dynamic,SViT,CROPR,zhou2023token,WyaJul_How_MICCAISAT2026}.
Further, expanding this work to other common token reduction approaches would help generalise these findings, for example with token merging or adaptive patch sizing. We also hope \ac{MCTP} informs future token reduction methods, for example in the choice of pretrained model, re-using pruning decisions across depths in coherent models to save compute, or estimating pruning aggression per class from its effective support. Lastly, approximating \ac{MCTP} at inference with uncertainty-driven approaches could help drive cheaper, higher quality pruning decisions.

\section{Conclusion}


In summary, we used \ac{MCTP} to investigate how pretrained models respond to pruning and how their token importance rankings develop through \ac{ViT} depth. We found that a model’s predictions under pruning depend on whether its high-norm sink tokens are kept, and that its \textbf{Coherent Aggregation} breaks when those predictions collapse. Removing DINOv3's prefix tokens collapsed its stability, while protecting only 5 of SigLIP 2's 441 tokens considerably raised it. Our results therefore suggest that token pruning should protect sink tokens: models trained with prefix tokens get this for free, while prefix-free models need their highest-norm tokens kept under pruning. Attention-based scores \cite{EViT, yang2025visionzip} and learned pruning modules \cite{rao2021dynamicvit, CROPR} already tend to keep these high-norm patch tokens implicitly; we hope our analysis of why these tokens matter guides future designs to protect sink tokens explicitly, such as registers or high-norm patch tokens, during token reduction.


\ifreviewsupp\section*{LLM Usage Statement}

Claude Code and Codex were used agentically for code generation, including the code that produces the figures and tables. ChatGPT was used for the generation of the example image in \cref{fig:method}. ChatGPT and Claude were used to assist with and tidy small sections of paper text and captions. Claude was also used interactively for project planning, related work analysis and paper framing, recording and organising the authors' ideas into notes rather than proposing the direction of the work.\fi

\ifreviewsupp\else
\section*{Acknowledgments}
This work was supported by the Engineering and Physical Sciences Research Council [grant number CS2324\_EPSRC\_1631720]. \\
The authors would like to acknowledge the use of the University of Oxford Advanced Research Computing (ARC) facility in carrying out this work. \\ \href{https://doi.org/10.5281/zenodo.22558}{\texttt{doi.org/10.5281/zenodo.22558}}\fi

{
    \small
    \bibliographystyle{ieeenat_fullname}
    \bibliography{main}
}

\clearpage
\ifreviewsupp\setcounter{page}{1}\fi
\maketitlesupplementary

\section{MCTP Sampling} \label{sec:MCTP-extra-details}
\subsection{Region Sampling}

\begin{figure}
    \centering
    \begin{tikzpicture}
        \node[inner sep=0pt, outer sep=0pt, anchor=south west] (grid) at (0,0)
            {\includegraphics[width=\columnwidth]{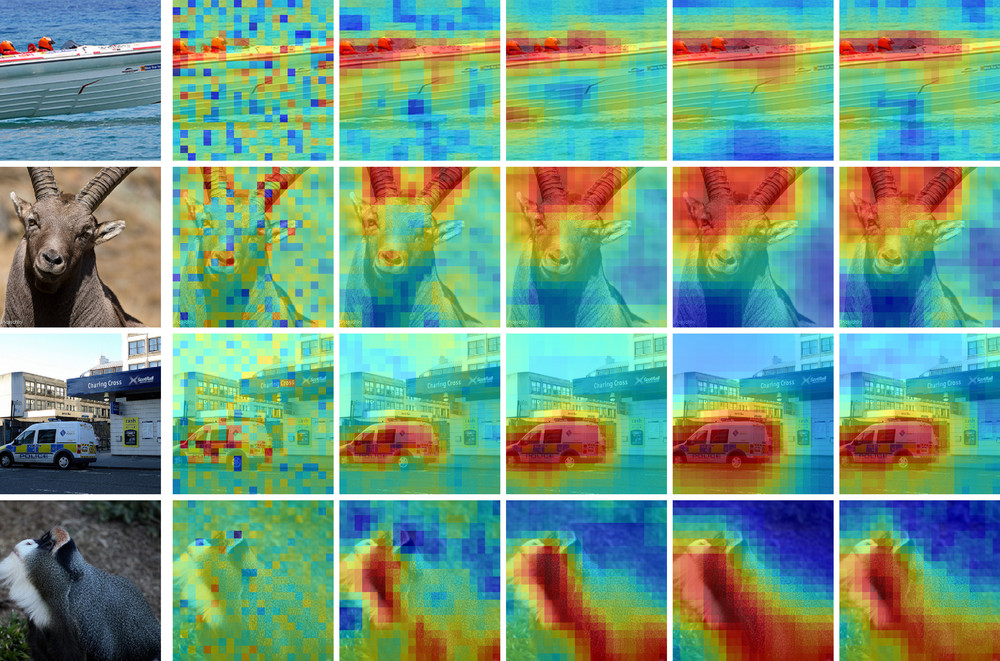}};

        \begin{scope}[
            x={(grid.south east)}, y={(grid.north west)},
            every node/.style={anchor=south, align=center, font=\scriptsize,
                               inner xsep=0pt, inner ysep=1pt}
        ]
        \node at (27/336,1)  {Input\\$|\Delta|\,(\mu {\pm} \sigma)$};
        \node at (85/336,1)  {Uniform\\$0.02{\pm}0.01$};
        \node at (141/336,1) {$3\times3$\\$0.07{\pm}0.03$};
        \node at (197/336,1) {$5\times5$\\$0.16{\pm}0.08$};
        \node at (253/336,1) {$7\times7$\\$0.21{\pm}0.08$};
        \node at (309/336,1) {$\{3,7\}$ mix\\$0.13{\pm}0.05$};
        \end{scope}
    \end{tikzpicture}
\caption{\textbf{MCTP sampler ablation.}
Token impact maps for DINOv3 B/16 at pruning depth $l{=}3$; uniform sampling uses $1\times1$ regions. Each heatmap is normalised independently. The headings give the per-image mean absolute impact $|\Delta_i|$, as mean $\pm$ SD over the first 1,000 ImageNet-1k validation images.}
    \label{fig:mctp-sampler-ablation}
\end{figure}

Uniformly sampled tokens change the prediction too little to separate one region from another: scattered tokens retain sufficient image context for the model to maintain the prediction regardless of the drawn $\tau$ regions, so their mean $|\Delta_i|$ is only $0.02$, against $0.07$--$0.21$ for region sampling (\cref{fig:mctp-sampler-ablation}). Therefore, we randomly sample larger contiguous, non-overlapping $n \times n$ regions subject to a nominal keep budget of $K{=}44$, 10\% of the 441 patch tokens. For the main paper results, we set $n\in\{3,7\}$.

More specifically, we sample the region size $n$ uniformly once per draw. First, we select the centre uniformly from all patch tokens and each subsequent centre from those that give a non-overlapping region, where regions extending outside of the image boundary are clipped. Each draw uses a budget of $\max(K,n^2)$ and stops when the remaining budget is below $n^2$ or no feasible centre remains, so before clipping a draw keeps four $3\times3$ regions (36 tokens) or one $7\times7$ region (49 tokens), and the $5\times5$ sampler in \cref{fig:mctp-sampler-ablation} keeps one (25 tokens). With clipping, the $\{3,7\}$ sampler keeps 40.2 patch tokens per draw on average (16--49).

\begin{figure}
    \centering
    \begin{tikzpicture}
        \node[inner sep=0pt, outer sep=0pt, anchor=south west] (grid) at (0,0)
            {\includegraphics[width=\columnwidth]{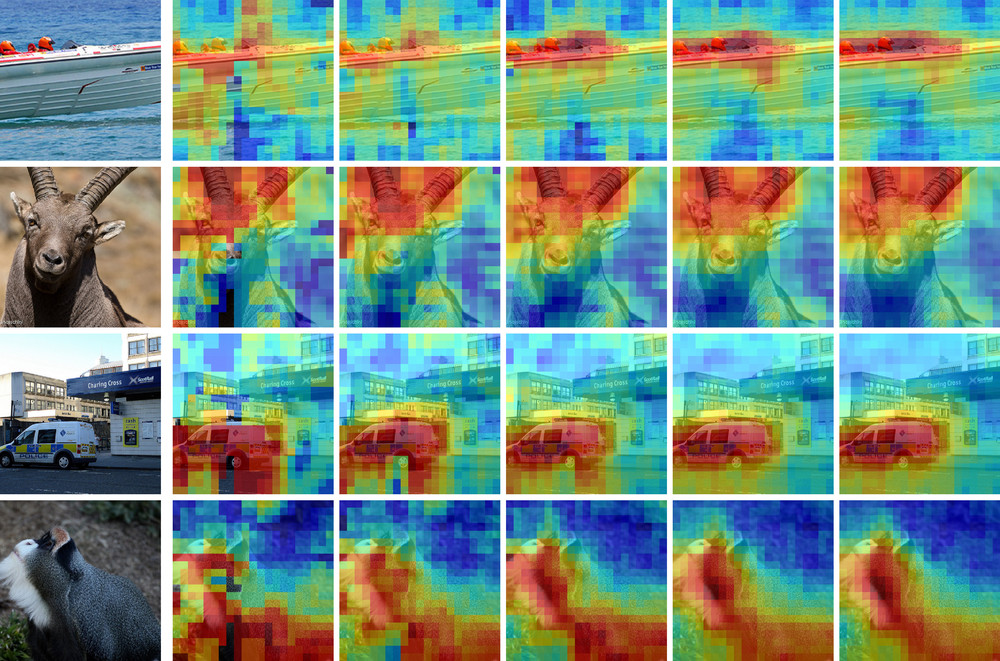}};

        \begin{scope}[
            x={(grid.south east)}, y={(grid.north west)},
            every node/.style={anchor=south, align=center, font=\scriptsize,
                               inner xsep=0pt, inner ysep=1pt}
        ]
            \node at (27/336,1)  {Input\\\phantom{$m=0.00$}};
            \node at (85/336,1)  {$M=50$\\$m=0.52$};
            \node at (141/336,1) {$M=100$\\$m=0.42$};
            \node at (197/336,1) {$M=200$\\$m=0.36$};
            \node at (253/336,1) {$M=500$\\$m=0.33$};
            \node at (309/336,1) {$M=1{,}000$\\$m=0.31$};
        \end{scope}
    \end{tikzpicture}
\caption{\textbf{Effect of the number of draws.}
Token impact maps on the images of \cref{fig:mctp-sampler-ablation}, using the first $M$ draws of one 1,000-draw run. At $M{=}50$, patches that have no impact estimate are left transparent and excluded from $m$. The headings give the mean off-box mass $m$ (\cref{eq:off-box}) over the 784 eligible images.}
    \label{fig:mctp-draw-count}
\end{figure}

\begin{figure}[htbp]
    \centering
    \includegraphics[width=\columnwidth]{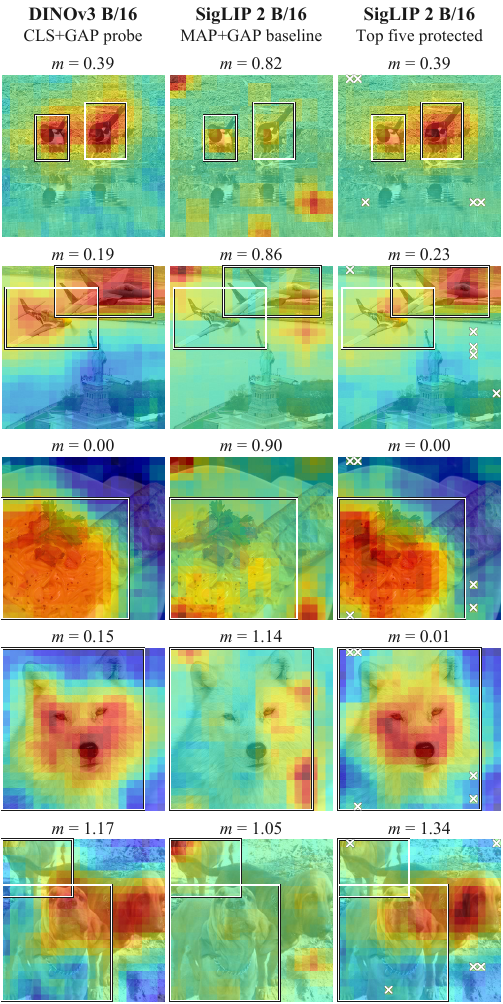}
    \caption{\textbf{Off-box impact on ImageNet-1k.} Signed token impacts at pruning depth \(l{=}6\) are overlaid on ImageNet-1k, with ground-truth box outlines and white crosses marking the five protected tokens in the right column. The headings give each image's area-normalised off-box mass \(m\), with \(m{=}1\) for uniform positive impact. In the bottom row, DINOv3 assigns strong positive impact to the rightmost dog's head, which lies outside both annotated boxes.}
    \label{fig:off-box-mass-examples}
\end{figure}

Over the 784 eligible images, those among the first 1,000 whose unpruned prediction is correct, the region size minimally changes \ac{MCTP} object alignment, with off-box mass $m$ (\cref{eq:off-box} and examples in \cref{fig:off-box-mass-examples}) varying by under 0.1 between region sizes. We omit uniform sampling here: $m$ is scale-free, so with near-zero impacts its value reflects noise.

\subsection{Number of Draws} \label{sec:mc-sample-size}

\Cref{fig:mctp-draw-count} shows the $\{3,7\}$ sampler on the same four images, using the first $M\in\{50,100,200,500,1{,}000\}$ draws of a single 1,000-draw run per image. At $M{=}50$, an average of 11 patches per image are not kept in any of the 50 draws and so have no impact estimate. Off-box mass settles as $M$ grows, changing by only 0.016 over the last 500 draws, so we use $M{=}1{,}000$ throughout.

\subsection{Keep Budget Sweep} \label{sec:k_sweep}

We repeat \ac{MCTP} at $l{=}3$ and $l{=}6$ on the first 1,000 validation images with keep budgets $K$ of 10, 25, 50 and 75\% (approximately 44, 110, 220 and 331 patch tokens), shown at $l{=}3$ in \cref{fig:mctp-keep-budget-sweep}. On average, this works out to 9.1, 21.4, 46.1 and 66.6\% of tokens actually kept because the keep regions are clipped at the border and a draw stops once no further region fits. For DINOv3, the magnitude of token impact drops tenfold between the 10\% and 75\% keep budgets because this directly causes the stability to rise. Consequently, the maps lose focus on the object as the off-box mass rises from 0.31 to 0.64, yet the effective support stays roughly the same. The opposite happens for SigLIP 2, as the larger keep budget lets more of its predictions survive, giving more tokens a small positive impact while the peaks stay in the same corners and the maps remain roughly uniform with respect to the object (off-box mass 1.14--1.17 at every budget). 

We additionally considered KL divergence between the unpruned and pruned predictive distributions. However, at the low keep budgets used to estimate $\Delta_i$, pruning induces a large distributional shift for most draws. We therefore use prediction stability on the reference class $\hat{y}$ as a more direct measure of its operational impact.

\subsection{Multi-seed Comparison} \label{sec:multi-seed}

\Cref{tab:multi-seed} compares five MCTP seeds. DINOv3 has an inter-seed Spearman correlation of 0.93 at every depth, close to its cross-depth correlation of 0.90, so its maps agree almost as well across depths as between seeds at the same depth. SigLIP 2's inter-seed correlation of 0.56--0.88 is far above its cross-depth correlation of 0.28, so its low \textbf{Coherent Aggregation} is not caused by Monte Carlo noise. For SigLIP 2, protecting the five highest-norm patch tokens at each depth reduces off-box mass and increases inter-seed agreement at all four depths.

\begin{table}[t]
\caption{\textbf{Multi-seed comparison.} DINOv3 and SigLIP 2 B/16
over five MCTP seeds. Entries are mean $\pm$ sample SD:
across five seed means for off-box mass $m$
(\cref{eq:off-box}; uniform impact gives $m{=}1$),
and across ten image-averaged seed-pair correlations for
signed inter-seed $\rho$. Pairwise SD is descriptive because
pairs share seeds. Protected tokens are excluded from both metrics.}
\centering
\footnotesize
\setlength{\tabcolsep}{2pt}
\begin{tabular}{@{}lcccc@{}}
\toprule
$l$ & 0 & 3 & 6 & 9 \\
\midrule
\multicolumn{5}{@{}l}{\textit{Off-box mass $m\downarrow$}} \\
DINOv3
 & $0.30 \pm 0.01$ & $0.31 \pm 0.01$
 & $0.31 \pm 0.00$ & $0.31 \pm 0.00$ \\
SigLIP 2
 & $0.97 \pm 0.02$ & $1.17 \pm 0.03$
 & $0.98 \pm 0.02$ & $0.87 \pm 0.01$ \\
\quad + top five
 & $0.92 \pm 0.01$ & $0.65 \pm 0.01$
 & $0.36 \pm 0.00$ & $0.54 \pm 0.01$ \\
\midrule
\multicolumn{5}{@{}l}{\textit{Inter-seed $\rho\uparrow$}} \\
DINOv3
 & $0.93 \pm 0.00$ & $0.93 \pm 0.00$
 & $0.93 \pm 0.00$ & $0.93 \pm 0.00$ \\
SigLIP 2
 & $0.60 \pm 0.02$ & $0.56 \pm 0.02$
 & $0.68 \pm 0.01$ & $0.88 \pm 0.00$ \\
\quad + top five
 & $0.61 \pm 0.01$ & $0.78 \pm 0.00$
 & $0.91 \pm 0.00$ & $0.95 \pm 0.00$ \\
\bottomrule
\end{tabular}
\label{tab:multi-seed}
\end{table}

\ifreviewsupp\else\begin{figure*}
    \centering
    \includegraphics[width=0.95\textwidth]{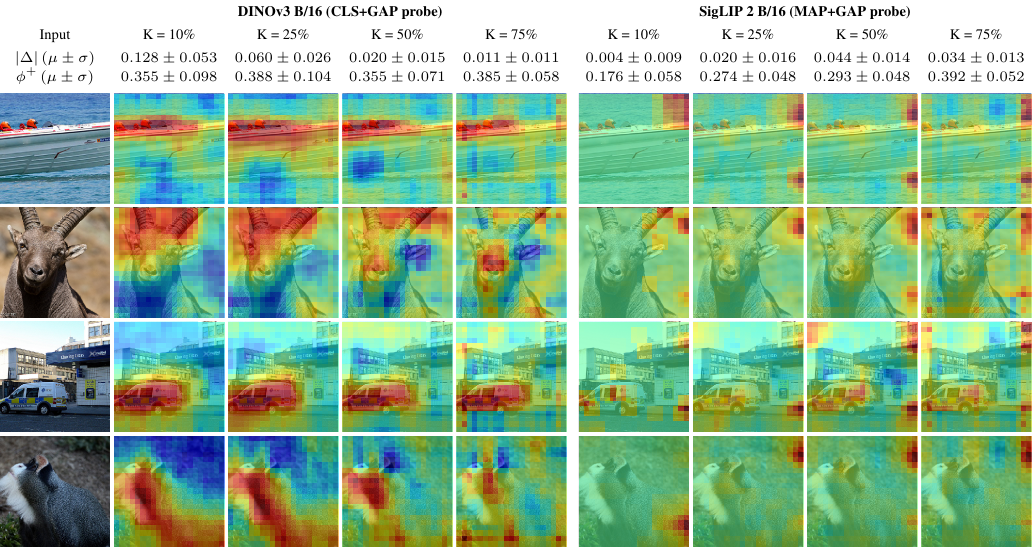}
\caption{\textbf{MCTP keep-budget sweep.}
    Token impact maps for DINOv3 B/16 (left) and SigLIP~2 B/16 (right) on the images of \cref{fig:mctp-sampler-ablation} at pruning depth $l{=}3$. The headings give the per-image mean absolute impact $|\Delta_i|$ and effective support $\phi^+$ (\cref{eq:effective-support}), each as mean $\pm$ SD over the first 1,000 ImageNet-1k validation images.}
    \label{fig:mctp-keep-budget-sweep}
\end{figure*}
\fi

\subsection{Efficiency} \label{sec:mctp-efficiency}

As in \cref{sec:MCTP-base}, \ac{MCTP} runs blocks $B_0\ldots B_{l{-}1}$ once per image and reuses that cache for every draw, so blocks $B_l\ldots B_{L{-}1}$ run on only the prefix tokens and the 10\% of patch tokens kept in $\tau$. At $l{=}6$, a draw costs 1.9 GFLOPs versus 41.8 GFLOPs for a full forward pass of DINOv3 B/16 at $336\times336$. Therefore, when $M{=}1{,}000$, this costs roughly the same as 47 forward passes, while occluding the input or attention-masking at each block would cost $1{,}000$. This is approximately 93 and 24 forward passes at $l{=}0$ and $l{=}9$ respectively. On two NVIDIA L40S GPUs with BF16 autocast, $l{=}6$ takes approximately 0.037 s per image, and all four depths over the 50,000 validation images take about two and a half hours.

\section{Extended Analysis}\label{sec:extended-analysis}

This section gives the full results for the six further models summarised in \cref{tab:backbone-summary}, run with the same \ac{MCTP} settings as \cref{sec:coherent-aggregation}: DeiT \cite{touvron2021training} (its official fine-tuned checkpoint, therefore not linear probed), LingBot-Vision \cite{fu2026vision}, EVA-02 \cite{eva02} (linear probed from its ImageNet-22k checkpoint for fairness), TIPSv2 \cite{tips_v2_paper}, and DINOv3 S/16 and L/16 \cite{simeoni2025dinov3} (\cref{sec:model-capacity}). As in the main paper, they span supervised classification, self-distillation, masked image modelling and image-text pretraining. We present Spearman correlations and top-10\% overlap (\cref{fig:supp-cross-depth-stability}), effective support distributions (\cref{fig:effective-support-distributions}), effective support against stability (\ifreviewsupp\cref{fig:depth-wise-siglip-effective-support,fig:effective-support-lingbot-deit,fig:effective-support-eva-tips,fig:effective-support-small-large}\else\cref{fig:depth-wise-siglip-effective-support,fig:effective-support-lingbot-deit}\fi) and token impact heatmaps (\cref{fig:futher_token_impacts_dinov3_siglip2,fig:futher_token_impacts_lingbot_deit,fig:futher_token_impacts_eva_tips,fig:futher_token_impacts_dinov3_siglip2_interventional}). For every model, the correlation falls as the two depths get further apart, but even the first and last depths still correlate at 0.62--0.86, against SigLIP 2's average of 0.28.

\ifreviewsupp\begin{figure}
    \centering
    \includegraphics[width=\linewidth, trim=0bp 13bp 0bp 0bp, clip]{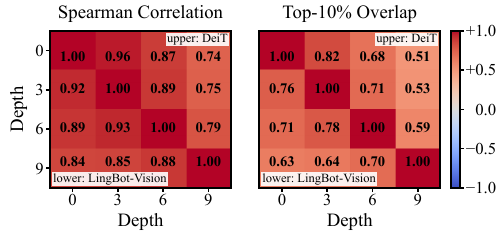}\\[2pt]
    \includegraphics[width=\linewidth, trim=0bp 13bp 0bp 11.5bp, clip]{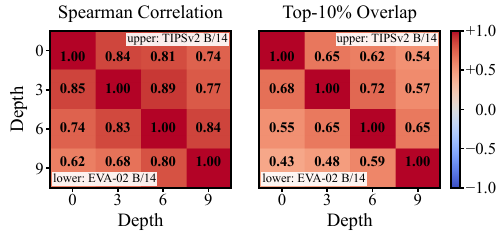}\\
    \includegraphics[width=\linewidth, trim=0bp 13bp 0bp 11.5bp, clip]{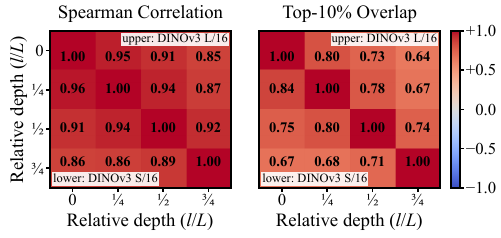}
    \caption{\textbf{Cross-depth consistency.} As in \cref{fig:cross-depth-stability-spearman-main}, for LingBot-Vision \cite{fu2026vision} and DeiT \cite{touvron2021training} (top), EVA-02 \cite{eva02} and TIPSv2 \cite{tips_v2_paper} (middle), and DINOv3 S/16 and L/16 \cite{simeoni2025dinov3} (bottom). The bottom row uses relative depth $l/L$, with $L{=}12$ for S/16 and $L{=}24$ for L/16.}
    \label{fig:supp-cross-depth-stability}
\end{figure}

\begin{figure}
    \centering
    \includegraphics[width=\linewidth, trim=0 12 0 0, clip]{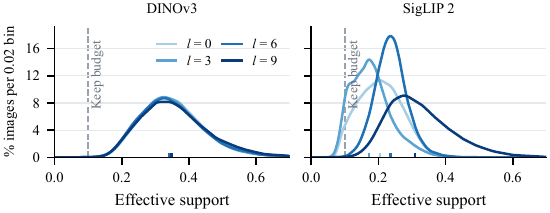}\\[2pt]
    \includegraphics[width=\linewidth, trim=0 12 0 0, clip]{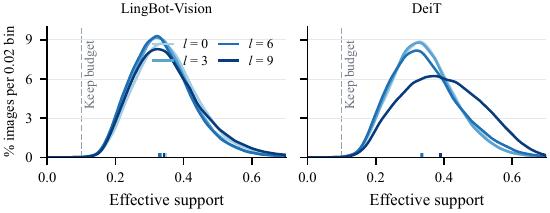}\\[2pt]
    \includegraphics[width=\linewidth, trim=0 12 0 0, clip]{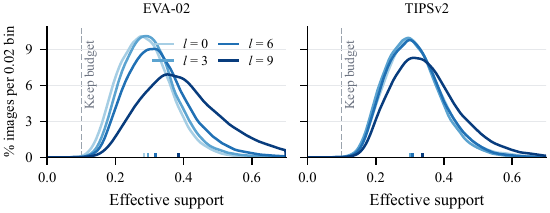}
    \includegraphics[width=\linewidth]{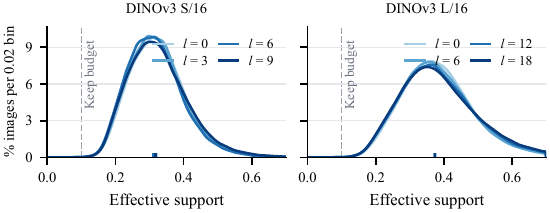}
\caption{\textbf{Effective support distributions.} Distribution of effective support $\phi^+$ (\cref{eq:effective-support}) over the ImageNet-1k validation set at each pruning depth ($l$), for DINOv3 \cite{simeoni2025dinov3}, SigLIP 2 \cite{tschannen2025siglip}, LingBot-Vision \cite{fu2026vision}, DeiT \cite{touvron2021training}, EVA-02 \cite{eva02} and TIPSv2 \cite{tips_v2_paper}, and DINOv3 S/16 and L/16.}
\label{fig:effective-support-distributions}
\end{figure}

\newcommand{\depthpanel}[4][]{%
  \begin{tikzpicture}
    \node[inner sep=0pt] (img) {\includegraphics[scale=0.85,#1]{#4}};
    \node[anchor=east, rectangle, rounded corners=1.5pt, draw=black!25, fill=black!4,
          line width=0.4pt, inner sep=1.5pt, text=black!80, font=\scriptsize]
      at ([shift={(-6.2bp,-#3bp)}]img.north east) {$l = #2$};
  \end{tikzpicture}}

\ifreviewsupp
\begin{figure*}[p]
    \centering
        \depthpanel[trim=0 12 0 0, clip]{0}{80.8}{figs/images/Analysis_2/supplementary/classwise_effective_support_vs_prediction_stability_depth0.pdf}\hspace{0.03\linewidth}%
    \depthpanel[trim=11 12 0 0, clip]{3}{80.9}{figs/images/Analysis_2/supplementary/classwise_effective_support_vs_prediction_stability_depth3.pdf}\\[2pt]
    \depthpanel{6}{82.4}{figs/images/Analysis_2/supplementary/classwise_effective_support_vs_prediction_stability_depth6.pdf}\hspace{0.03\linewidth}%
    \depthpanel[trim=11 0 0 0, clip]{9}{84.1}{figs/images/Analysis_2/supplementary/classwise_effective_support_vs_prediction_stability_depth9.pdf}
    \caption{\textbf{Effective support across depths.} As in \cref{fig:effective-support-stab-main}, for SigLIP 2 \cite{tschannen2025siglip} at $l{=}0,3$ (top) and $l{=}6,9$ (bottom), without the purple below-trend classes. Screwdriver and jigsaw puzzle are marked in every panel so they can be followed across depths and models. Pearson correlations are 0.61, 0.46, 0.74 and 0.96 at $l{=}0,3,6,9$.}
    \label{fig:depth-wise-siglip-effective-support}

    \vspace{8pt}
    \includegraphics[width=0.78\linewidth]{figs/images/Analysis_2/supplementary/effective_support_support_stability_lingbot_deit.pdf}
\caption{\textbf{Further effective support results.} As in \cref{fig:effective-support-stab-main}, for LingBot-Vision \cite{fu2026vision} (left) and DeiT \cite{touvron2021training} (right) at $l{=}3$.}
\label{fig:effective-support-lingbot-deit}

    \vspace{8pt}
    \includegraphics[width=0.78\linewidth]{figs/images/Analysis_2/supplementary/effective_support_support_stability_eva_tips.pdf}
\caption{\textbf{Further effective support results.} As in \cref{fig:effective-support-stab-main}, for EVA-02 \cite{eva02} (left) and TIPSv2 \cite{tips_v2_paper} (right) at $l{=}3$.}
    \label{fig:effective-support-eva-tips}
\end{figure*}

\begin{figure*}
    \centering
    \includegraphics[width=0.78\linewidth]{figs/images/Analysis_2/supplementary/effective_support_support_stability_dinov3_small_large.pdf}
\caption{\textbf{Further effective support results.} As in \cref{fig:effective-support-stab-main}, for DINOv3 \cite{simeoni2025dinov3} S/16 at $l{=}3$ (left) and L/16 at $l{=}6$ (right), the same relative depth.}
    \label{fig:effective-support-small-large}
\end{figure*}

\else
\begin{figure*}[p]
    \centering
    \includegraphics{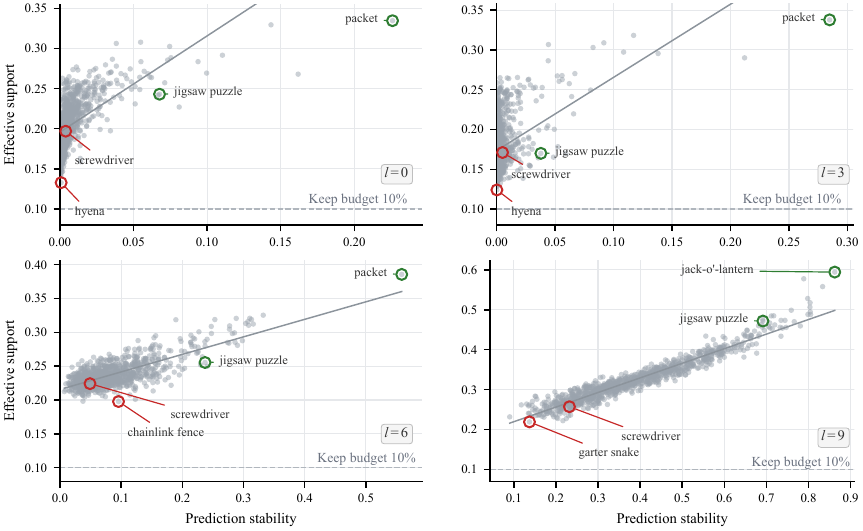}
    \caption{\textbf{Effective support across depths.} As in \cref{fig:effective-support-stab-main}, for SigLIP 2 \cite{tschannen2025siglip} at each pruning depth, without the purple below-trend classes. Screwdriver and jigsaw puzzle are marked in every panel so they can be followed across depths and models. Pearson correlations are 0.61, 0.46, 0.74 and 0.96 at $l{=}0,3,6,9$.}
    \label{fig:depth-wise-siglip-effective-support}

    \vspace{8pt}
    \includegraphics{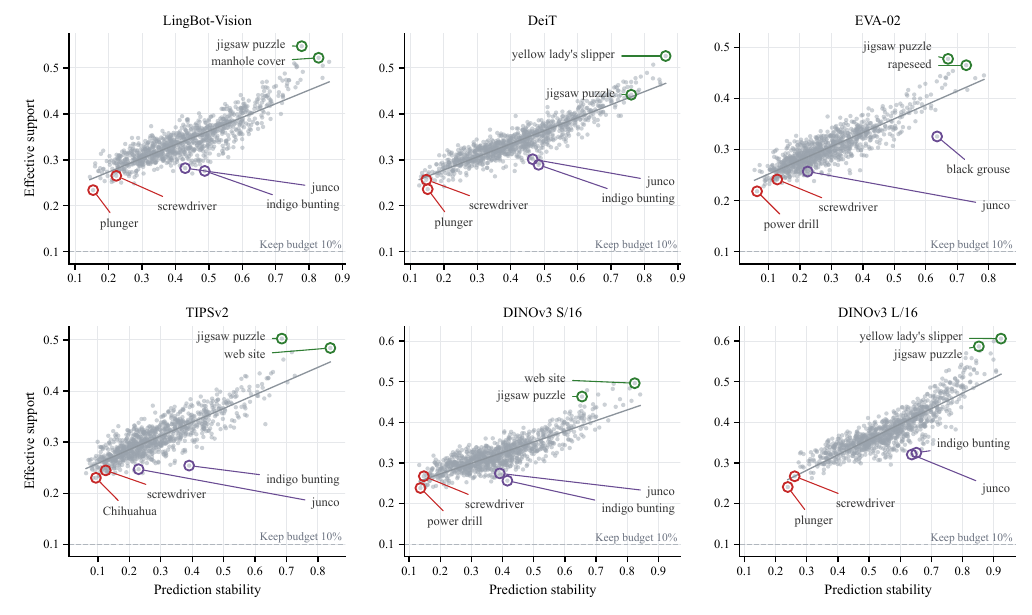}
    \caption{\textbf{Further effective support results.} As in \cref{fig:effective-support-stab-main}, for LingBot-Vision \cite{fu2026vision}, DeiT \cite{touvron2021training}, EVA-02 \cite{eva02}, TIPSv2 \cite{tips_v2_paper} and DINOv3 \cite{simeoni2025dinov3} S/16 and L/16, all at $l{=}3$ except L/16 at $l{=}6$, the same relative depth.}
    \label{fig:effective-support-lingbot-deit}\label{fig:effective-support-eva-tips}\label{fig:effective-support-small-large}
\end{figure*}
\fi
\fi
\begin{table}
\caption{\textbf{Prediction stability and rank agreement across pruning depths.}
Prediction stability $\hat{s}$ (\cref{eq:prediction-stability}) at each pruning depth $l$, and mean Spearman $\rho$, over the ImageNet-1k validation set. $\hat{s}$ is measured against the original, unmodified model's unpruned top-1. $\rho$ averages per-image rank correlations of signed token impacts over the six depth pairs, using only positions unprotected at all four depths. DINOv3 L/16 uses $l \in \{0,6,12,18\}$.}
\centering
\footnotesize
\setlength{\tabcolsep}{4pt}
\begin{tabular}{lccccc}
\toprule
 & \multicolumn{4}{c}{$\hat{s}$ (\%) $\uparrow$} & \\
\cmidrule(lr){2-5}
Model & $l{=}0$ & $l{=}3$ & $l{=}6$ & $l{=}9$ & $\rho \uparrow$ \\
\midrule
DINOv3 B/16        & 48.71 & 48.03 & 52.03 & 58.05 & 0.90 \\
LingBot-Vision B/16 & 47.46 & 43.12 & 45.90 & 55.73 & 0.89 \\
DeiT B/16          & 42.30 & 42.41 & 50.35 & 68.94 & 0.83 \\
TIPSv2 B/14        & 34.54 & 28.59 & 36.54 & 54.44 & 0.82 \\
EVA-02 B/14        & 17.02 & 28.76 & 43.23 & 62.82 & 0.75 \\
SigLIP 2 B/16      &  0.86 &  0.58 &  8.69 & 38.78 & 0.28 \\
\midrule
DINOv3 S/16        & 40.72 & 38.79 & 39.49 & 46.39 & 0.90 \\
DINOv3 L/16        & 55.71 & 56.60 & 60.85 & 64.54 & 0.91 \\
\midrule
DINOv3 B/16 (GAP)           & 43.88 & 45.18 & 49.36 & 55.20 & 0.90 \\
\quad Prefix removed        &  0.21 &  0.29 &  3.86 & 36.02 & 0.34 \\
SigLIP 2 B/16               &  0.86 &  0.58 &  8.69 & 38.78 & 0.28 \\
\quad Top five protected    &  2.69 & 10.61 & 36.60 & 65.95 & 0.40 \\
\quad Test-time registers   &  0.42 &  0.74 & 30.86 & 65.88 & 0.21 \\
\quad Random five protected &  0.93 &  0.74 & 11.71 & 43.54 & 0.28 \\
\quad Empty register        &  0.49 &  0.62 &  8.74 & 39.34 & 0.28 \\
\bottomrule
\end{tabular}
\label{tab:stability-depth}
\end{table}

\subsection{Model Capacity Scaling}\label{sec:model-capacity}
Following the rest of the extended analysis, we briefly verify our approach on DINOv3 S/16 and L/16 to validate a consistency against both pretraining schemes and model sizes. Both models are distilled from the same 7B parameter teacher, therefore consistency in this case is a property of the pretraining recipe and not capacity. As all other experiments use 12-block transformers, we prune L/16 at the same relative depths, $l \in \{0, 6, 12, 18\}$. The models remain highly consistent when scaling capacity, with the main difference resulting from the greater stability for larger models (\cref{tab:stability-depth}).

\subsection{Global Average Token Impact} \label{sec:global-average-impact}
\ifreviewsupp\else\input{figs/supp_global_impact}\fi
We calculate the average impact over the whole of the ImageNet-1k validation set (\cref{fig:global-impact}). DINOv3's global impact conforms with standard image classification centre bias, while SigLIP 2 has no central peak and its positive impact sits along the image edges and corners.

\subsection{Class Ranking Consistency}
\label{sec:class-ranking}
To test how much prediction stability depends on the model rather than the data, we rank the 1,000 ImageNet-1k classes by $\hat{s}(k)$ and compare these rankings between models using Spearman correlation. At $l{=}3$ ($l{=}6$ for L/16), the 7 backbones other than SigLIP 2 rank the classes by $\hat{s}(k)$ similarly, with correlations of 0.70--0.96 between every pair of models (0.92--0.96 between the 3 DINOv3 sizes), and their rankings by effective support $\phi^{+}(k)$ agree just as closely (0.76--0.96). Which classes keep their prediction is therefore largely set by the images: at a 10\% keep budget, $\hat{s}(k)$ measures whether a small region of the image suffices to classify it, as well as how the model responds to losing the rest. SigLIP 2 is the exception: it keeps almost no predictions at $l{=}3$, and even at $l{=}9$, where 38.8\% of its draws keep the prediction, its rankings correlate with the other models at only 0.38--0.46 for $\hat{s}(k)$ and 0.32--0.38 for $\phi^{+}(k)$. Protecting its top five tokens lets us compare it at $l{=}3$, where these rise to 0.56--0.74 for $\hat{s}(k)$ and 0.43--0.62 for $\phi^{+}(k)$, suggesting that much of what sets SigLIP 2 apart comes from losing its sinks rather than from the images.

\subsection{Sink Depth-wise Overlap and Formation}
\ifreviewsupp\else\begin{figure}
    \centering
    \includegraphics[width=\linewidth]{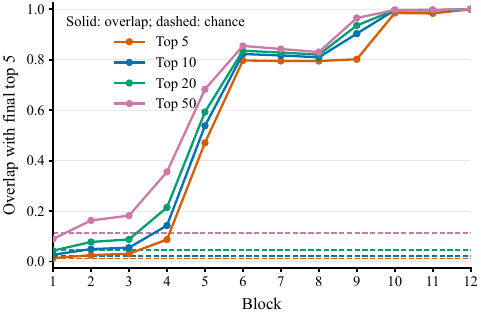}
\caption{\textbf{Highest-norm token overlap.} Share of the final block's five highest-norm patch tokens that are among the $k$ highest-norm tokens at each block, for unpruned SigLIP~2 B/16 over the ImageNet-1k validation set.}

    \label{fig:topk-norm-overlap}
\end{figure}
\begin{table}[t]
\caption{\textbf{Sink formation under pruning.} SigLIP 2 B/16 on 1,000 ImageNet-1k validation images, using the first 50 of the 1,000 draws per image. A sink is a kept patch token whose $\ell_2$ norm exceeds $5\times$ the median kept-patch norm at that block. \textit{Sink}: share of draws (\%) with a sink. \textit{Prot.}: of those draws, the share (\%) in which a protected token is a sink.}
\centering
\footnotesize
\setlength{\tabcolsep}{3.5pt}
\begin{tabular}{llcccccc}
\toprule
 & & \multicolumn{2}{c}{Block 6} & \multicolumn{2}{c}{Block 9} & \multicolumn{2}{c}{Block 12} \\
\cmidrule(lr){3-4}\cmidrule(lr){5-6}\cmidrule(lr){7-8}
$l$ & Protection & Sink & Prot. & Sink & Prot. & Sink & Prot. \\
\midrule
0 & None & 7.1 & -- & 3.5 & -- & 10.1 & -- \\
  & Random five & 9.2 & 6 & 5.7 & 5 & 11.6 & 8 \\
  & Top five & 8.9 & 49 & 5.0 & 61 & 32.8 & 89 \\
\midrule
3 & None & 14.0 & -- & 11.7 & -- & 15.8 & -- \\
  & Random five & 15.4 & 14 & 12.9 & 11 & 17.3 & 15 \\
  & Top five & 17.2 & 92 & 12.8 & 95 & 38.3 & 99 \\
\bottomrule
\end{tabular}
\label{tab:sink-formation}
\end{table}
\fi

\cref{fig:topk-norm-overlap} and \cref{tab:sink-formation} give the full results behind the sink formation analysis for SigLIP 2 in \cref{sec:coherent-causes}. To add to that discussion, \cref{fig:topk-norm-overlap} measures $|\operatorname{top}_k(l)\cap\operatorname{top}_5(12)|/5$, where $\operatorname{top}_k(l)$ is the set of $k$ patch tokens in $z^l$ with the highest $\ell_2$ norm, averaged over the validation set; random sets give $k/441$. This tests whether the tokens we protect at block 3 are the ones that become sinks in the unpruned model. To summarise our original discussion, the high-norm tokens develop by the 6th block, where blocks 4--6 see the largest increase in overlap with the final top five relative to the previous block. The five highest-norm tokens at block 3, which we protect when pruning at $l{=}3$, contain only 3.1\% of the final block’s top five (chance 1.1\%), against 80\% at block 6, and even the 50 highest-norm tokens at block 3 contain only about 18\%. This indicates that the highest-norm tokens at block 3 are separate from those that become sinks later in the model, and so does not explain why protecting the top five highest-norm tokens at $l{=}3$ improves stability (\cref{tab:stability-depth}). 

Therefore, \cref{tab:sink-formation} measures how often SigLIP 2 still forms a sink under pruning at blocks 6, 9 and 12, and whether it is one of the protected tokens. When the protected tokens are selected by highest-norm, a sink that forms is likely to be the protected token. However, the rate at which sinks form remains low across all interventions at $l{=}0$ (3.5--32.8\%) and increases slightly at $l{=}3$.

\ifreviewsupp\begin{table}[t]
\caption{\textbf{Sink formation under pruning.} SigLIP 2 B/16 on 1,000 ImageNet-1k validation images, using the first 50 of the 1,000 draws per image. A sink is a kept patch token whose $\ell_2$ norm exceeds $5\times$ the median kept-patch norm at that block. \textit{Sink}: share of draws (\%) with a sink. \textit{Prot.}: of those draws, the share (\%) in which a protected token is a sink.}
\centering
\footnotesize
\setlength{\tabcolsep}{3.5pt}
\begin{tabular}{llcccccc}
\toprule
 & & \multicolumn{2}{c}{Block 6} & \multicolumn{2}{c}{Block 9} & \multicolumn{2}{c}{Block 12} \\
\cmidrule(lr){3-4}\cmidrule(lr){5-6}\cmidrule(lr){7-8}
$l$ & Protection & Sink & Prot. & Sink & Prot. & Sink & Prot. \\
\midrule
0 & None & 7.1 & -- & 3.5 & -- & 10.1 & -- \\
  & Random five & 9.2 & 6 & 5.7 & 5 & 11.6 & 8 \\
  & Top five & 8.9 & 49 & 5.0 & 61 & 32.8 & 89 \\
\midrule
3 & None & 14.0 & -- & 11.7 & -- & 15.8 & -- \\
  & Random five & 15.4 & 14 & 12.9 & 11 & 17.3 & 15 \\
  & Top five & 17.2 & 92 & 12.8 & 95 & 38.3 & 99 \\
\bottomrule
\end{tabular}
\label{tab:sink-formation}
\end{table}
\fi

\ifreviewsupp
\begin{figure*}[t]
  \renewenvironment{figure}[1][]{}{}%
  \begin{minipage}[t]{0.48\linewidth}\vspace{0pt}\input{figs/supp_global_impact}\end{minipage}\hfill
  \begin{minipage}[t]{0.48\linewidth}\vspace{0pt}\input{figs/sink_overlap_formation}\end{minipage}
\end{figure*}
\fi

\section{Hyperparameters} \label{sec:hyperparameters}

In \cref{tab:model-hyperparameters,tab:linear-probe-hyperparameters,tab:mctp-hyperparameters} we list hyperparameters for the models, the linear probes and \ac{MCTP}. Furthermore, we provide the top-1 accuracy for our linear probes in \cref{tab:readout-ablation}.

\ifreviewsupp\else
\vspace*{\fill}
\begingroup
\let\origfigure\figure \let\endorigfigure\endfigure
\renewenvironment{figure}[1][]{\origfigure[h]}{\endorigfigure}
\begin{figure}
    \centering
    \includegraphics[width=\linewidth, trim=0bp 13bp 0bp 0bp, clip]{figs/images/Analysis_1/lingbot_vision_deit.pdf}\\[2pt]
    \includegraphics[width=\linewidth, trim=0bp 13bp 0bp 11.5bp, clip]{figs/images/Analysis_1/eva02_tipsv2.pdf}\\
    \includegraphics[width=\linewidth, trim=0bp 13bp 0bp 11.5bp, clip]{figs/images/Analysis_1/dinov3_small_large.pdf}
    \caption{\textbf{Cross-depth consistency.} As in \cref{fig:cross-depth-stability-spearman-main}, for LingBot-Vision \cite{fu2026vision} and DeiT \cite{touvron2021training} (top), EVA-02 \cite{eva02} and TIPSv2 \cite{tips_v2_paper} (middle), and DINOv3 S/16 and L/16 \cite{simeoni2025dinov3} (bottom). The bottom row uses relative depth $l/L$, with $L{=}12$ for S/16 and $L{=}24$ for L/16.}
    \label{fig:supp-cross-depth-stability}
\end{figure}

\begin{figure}
    \centering
    \includegraphics[width=\linewidth, trim=0 12 0 0, clip]{figs/images/Analysis_2/effective_support_distributions.pdf}\\[2pt]
    \includegraphics[width=\linewidth, trim=0 12 0 0, clip]{figs/images/Analysis_2/supplementary/effective_support_distributions_lingbot_deit.pdf}\\[2pt]
    \includegraphics[width=\linewidth, trim=0 12 0 0, clip]{figs/images/Analysis_2/supplementary/effective_support_distributions_eva_tips.pdf}
    \includegraphics[width=\linewidth]{figs/images/Analysis_2/supplementary/effective_support_distributions_dinov3_small_large.pdf}
\caption{\textbf{Effective support distributions.} Distribution of effective support $\phi^+$ (\cref{eq:effective-support}) over the ImageNet-1k validation set at each pruning depth ($l$), for DINOv3 \cite{simeoni2025dinov3}, SigLIP 2 \cite{tschannen2025siglip}, LingBot-Vision \cite{fu2026vision}, DeiT \cite{touvron2021training}, EVA-02 \cite{eva02} and TIPSv2 \cite{tips_v2_paper}, and DINOv3 S/16 and L/16.}
\label{fig:effective-support-distributions}
\end{figure}

\endgroup
\vspace*{\fill}

\fi

\input{figs/token_impacts_extended}

\ifreviewsupp
  \begin{figure*}
    \centering
    \includegraphics[width=0.95\textwidth]{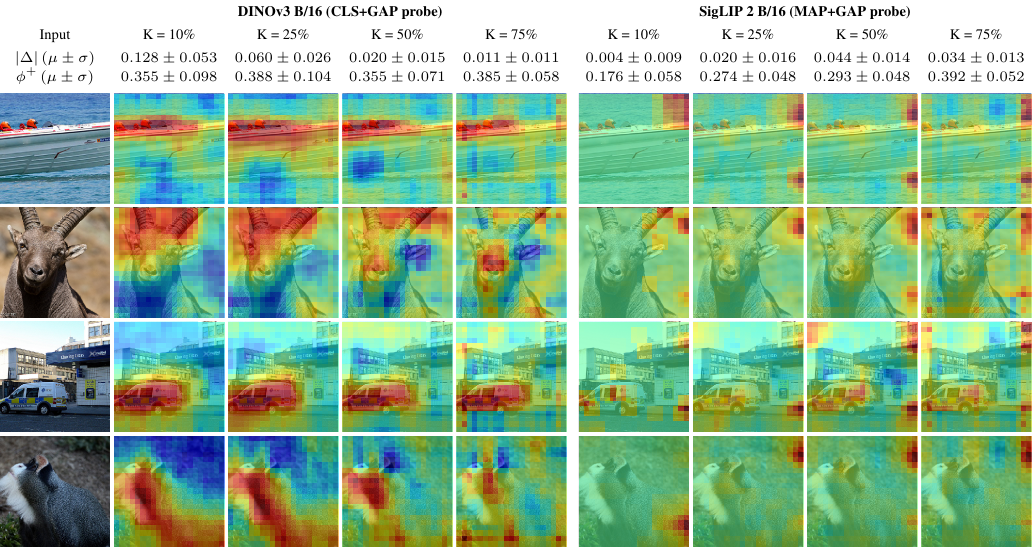}
\caption{\textbf{MCTP keep-budget sweep.}
    Token impact maps for DINOv3 B/16 (left) and SigLIP~2 B/16 (right) on the images of \cref{fig:mctp-sampler-ablation} at pruning depth $l{=}3$. The headings give the per-image mean absolute impact $|\Delta_i|$ and effective support $\phi^+$ (\cref{eq:effective-support}), each as mean $\pm$ SD over the first 1,000 ImageNet-1k validation images.}
    \label{fig:mctp-keep-budget-sweep}
\end{figure*}

\else
  \clearpage
\fi

\begin{table}[p]
\caption{\textbf{Readout ablation.} Top-1 (\%) of the unpruned linear probes for each readout at $336{\times}336$, over the ImageNet-1k validation set. Bold marks the best readout for each model. All MCTP results use CLS+GAP (\cref{eq:readout}), except DeiT, which keeps its fine-tuned classifier, and the DINOv3 prefix-removal runs, which use GAP; CLS+GAP is within 0.1 points of the best readout for every other model.}
\label{tab:readout-ablation}
\centering
\small
\begin{tabular}{lccc}
\toprule
Model & CLS & GAP & CLS+GAP \\
\midrule
DINOv3 B/16 & 85.1 & 81.4 & \textbf{85.3} \\
SigLIP 2 B/16\textsuperscript{$a$} & \textbf{85.0} & 82.9 & 84.9 \\
LingBot-Vision B/16 & 84.4 & 76.2 & \textbf{84.6} \\
DeiT B/16\textsuperscript{$b$} & \textbf{82.0} & -- & -- \\
EVA-02 B/14\textsuperscript{$c$} & 81.9 & \textbf{82.7} & \textbf{82.7} \\
TIPSv2 B/14\textsuperscript{$c$} & 82.0 & 80.9 & \textbf{82.7} \\
\midrule
DINOv3 S/16 & 80.6 & 72.2 & \textbf{80.9} \\
DINOv3 L/16 & 87.4 & 84.1 & \textbf{87.5} \\
\bottomrule
\end{tabular}
\par\smallskip
\begin{minipage}{0.98\linewidth}
\footnotesize
\textsuperscript{$a$}No CLS token; uses the multi-head attention pooling (MAP) output in its place.
\textsuperscript{$b$}Uses its original ImageNet-1k fine-tuned CLS classifier.
\textsuperscript{$c$}Patch projection resampled from 14 to 16 pixels.
\end{minipage}
\end{table}

\begin{table}[p]
\caption{\textbf{Backbones.} Pretrained models evaluated at $336\times336$
with 16-pixel patches.}
\label{tab:model-hyperparameters}
\centering
\small
\setlength{\tabcolsep}{3pt}
\begin{tabular*}{\columnwidth}{@{\extracolsep{\fill}}lll@{}}
\toprule
Model & Pretraining & Prefix tokens \\
 & objective & \\
\midrule
DINOv3 B/16 & Self-distillation & CLS + 4 registers \\
 & + MIM & \\
\multicolumn{3}{@{}l@{}}{\hspace{0.6em}{\scriptsize\texttt{vit\_base\_patch16\_dinov3.lvd1689m}}} \\
SigLIP~2 B/16 & Sigmoid & None \\
 & image--text + SSL & \\
\multicolumn{3}{@{}l@{}}{\hspace{0.6em}{\scriptsize\texttt{google/siglip2-base-patch16-384}}} \\
LingBot-Vision B/16 & Self-dist. and masked & CLS + 4 registers \\
 & boundary modelling & \\
\multicolumn{3}{@{}l@{}}{\hspace{0.6em}{\scriptsize\texttt{robbyant/lingbot-vision-vit-base}}} \\
DeiT B/16 & Supervised & CLS \\
 & classification & \\
\multicolumn{3}{@{}l@{}}{\hspace{0.6em}{\scriptsize\texttt{deit\_base\_patch16\_224.fb\_in1k}}} \\
EVA-02 B/14\textsuperscript{a} & MIM (EVA-CLIP & CLS \\
 & teacher) & \\
\multicolumn{3}{@{}l@{}}{\hspace{0.6em}{\scriptsize\texttt{eva02\_base\_patch14\_224.mim\_in22k}}} \\
TIPSv2 B/14\textsuperscript{a} & Image--text + & CLS + 1 auxiliary\textsuperscript{b} \\
 & patch distillation & \\
\multicolumn{3}{@{}l@{}}{\hspace{0.6em}{\scriptsize\texttt{google/tipsv2-b14}}} \\
\midrule
DINOv3 S/16 & Self-distillation & CLS + 4 registers \\
 & + MIM & \\
\multicolumn{3}{@{}l@{}}{\hspace{0.6em}{\scriptsize\texttt{vit\_small\_patch16\_dinov3.lvd1689m}}} \\
DINOv3 L/16\textsuperscript{c} & Self-distillation & CLS + 4 registers \\
 & + MIM & \\
\multicolumn{3}{@{}l@{}}{\hspace{0.6em}{\scriptsize\texttt{vit\_large\_patch16\_dinov3.lvd1689m}}} \\
\bottomrule
\end{tabular*}
\par\smallskip
\begin{minipage}{\columnwidth}
\footnotesize
\textsuperscript{a}The pretrained $14\times14$ patch projections (EVA-02, TIPSv2) are
resampled to $16\times16$.
\textsuperscript{b}TIPSv2's auxiliary token is a second class token; the CLS readout uses the first.
\textsuperscript{c}DINOv3 L/16 has 24 blocks, and every other model has 12.
\end{minipage}
\end{table}

\begin{table}[p]
\caption{\textbf{Linear-probe settings.} Shared training settings for the fitted probes.}
\label{tab:linear-probe-hyperparameters}
\centering
\small
\setlength{\tabcolsep}{3pt}
\begin{tabular*}{\columnwidth}{@{\extracolsep{\fill}}ll@{}}
\toprule
Hyperparameter & Value \\
\midrule
Optimiser & SGD; momentum $0.9$ \\
Weight decay & $0$ \\
Loss & Cross-entropy \\
Effective batch size & 1,024 \\
Training steps & $10\times1{,}250=12{,}500$ \\
Base LR grid & $\{1,2,5\}\!\times\!10^{j}$, \\
 & $j\in\{-5,-4,-3,-2\}$; $10^{-1}$ \\
LR selection split & ImageNet-1k validation set \\
LR schedule & Cosine to zero; 0 warm-up steps \\
Training transforms & Random resized crop to 336; \\
 & horizontal flip ($p=0.5$) \\
Evaluation transforms & Resize short side to 336; \\
 & centre crop to 336 \\
Normalisation & ImageNet\textsuperscript{a}; SigLIP~2: \\
 & $(x-0.5)/0.5$; TIPSv2: $x$ \\
Random seed & $42$ \\
\bottomrule
\end{tabular*}
\par\smallskip
\begin{minipage}{\columnwidth}
\footnotesize
\textsuperscript{a}Mean $(0.485,0.456,0.406)$; std $(0.229,0.224,0.225)$.
Resizing is bicubic. Base learning rates are multiplied by
$B_{\mathrm{eff}}/256=4$ and selected per readout using final-epoch
validation top-1.
\end{minipage}
\end{table}

\begin{table}[p]
\caption{\textbf{MCTP settings.} Sampling and evaluation settings.}
\label{tab:mctp-hyperparameters}
\centering
\small
\setlength{\tabcolsep}{3pt}
\begin{tabular*}{\columnwidth}{@{\extracolsep{\fill}}ll@{}}
\toprule
Hyperparameter & Value \\
\midrule
Patch tokens, $N$ & $21\times21=441$ \\
Keep budget, $K$ & $10\%$ of $N$ (44 tokens)\textsuperscript{a} \\
Draws, $M$ & 1,000 \\
Region side, $n$ & $\{3,7\}$, sampled uniformly per draw \\
Pruning depths, $l$ & $\{0,3,6,9\}$; $\{0,6,12,18\}$ \\
 & for DINOv3 L/16 \\
Random seed & $42$ \\
Precision & BF16 draws; FP32 unpruned reference \\
Hardware & Two NVIDIA L40S GPUs (most runs) \\
\bottomrule
\end{tabular*}
\par\smallskip
\begin{minipage}{\columnwidth}
\footnotesize
\textsuperscript{a}Nominal: draws keep 16--49 patch tokens, 40.2 on average.
The same 1,000 token selections are used for every image, model and depth;
the five-seed comparison uses seeds 42--46.
\end{minipage}
\end{table}

\end{document}